\documentclass[10pt,twocolumn,letterpaper]{article}

\usepackage[pagenumbers]{wacv} 
\usepackage{multirow}
\usepackage{float}
\usepackage{caption}
\usepackage{xcolor}
\usepackage{enumitem}

\usepackage{booktabs}
\usepackage{array}
\usepackage{etoolbox}

\AtBeginDocument{%
  \setlength{\abovedisplayskip}{3pt plus 1pt minus 1pt}%
  \setlength{\belowdisplayskip}{3pt plus 1pt minus 1pt}%
  \setlength{\abovedisplayshortskip}{2pt plus 1pt minus 1pt}%
  \setlength{\belowdisplayshortskip}{2pt plus 1pt minus 1pt}%
}
\definecolor{wacvblue}{rgb}{0.21,0.49,0.74}
\usepackage[pagebackref,breaklinks,colorlinks,allcolors=wacvblue]{hyperref}

\def\wacvPaperID{386} 
\def\confName{WACV}
\def\confYear{2027}
\title{\textit{PLS-GS}: Probabilistic Lidar Scaffolds for\\%
Compact, Surface-Aligned Gaussian Splatting}

\author{
Krishnan Harikumar$^{1}$\hspace{0.5em}
Alejandro Fontan$^{1}$\hspace{0.5em}
Thierry Peynot$^{1}$\hspace{0.5em}
Emili Hern\'andez$^{2}$\hspace{0.5em}
Niko S\"underhauf$^{1}$ \\
$^1$QUT Centre for Robotics \quad
$^2$Emesent \\
{\tt\small bharathkrishnan.harikumar@hdr.qut.edu.au} \\
{\tt\small \{alejandro.fontan, t.peynot, niko.suenderhauf\}@qut.edu.au, emili.hernandez@emesent.com}
}

\begin{document}
\AddToShipoutPicture*{\AtTextUpperLeft{\put(0,22){\parbox{\textwidth}{\centering\scriptsize
PREPRINT VERSION OF A 2027 IEEE/CVF WINTER CONFERENCE ON APPLICATIONS OF COMPUTER VISION (WACV) PAPER; FINAL VERSION TO APPEAR ON IEEE XPLORE}}}}
\AddToShipoutPicture*{\AtTextLowerLeft{\put(0,-55){\parbox{\textwidth}{\scriptsize
\copyright~2026 IEEE. Personal use of this material is permitted. Permission from IEEE must be obtained for all other uses, in any current or future media, including reprinting/republishing this material for advertising or promotional purposes, creating new collective works, for resale or redistribution to servers or lists, or reuse of any copyrighted component of this work in other works.}}}}
\maketitle

\begin{figure*}[!t]
    \centering
    \includegraphics[width=\textwidth]{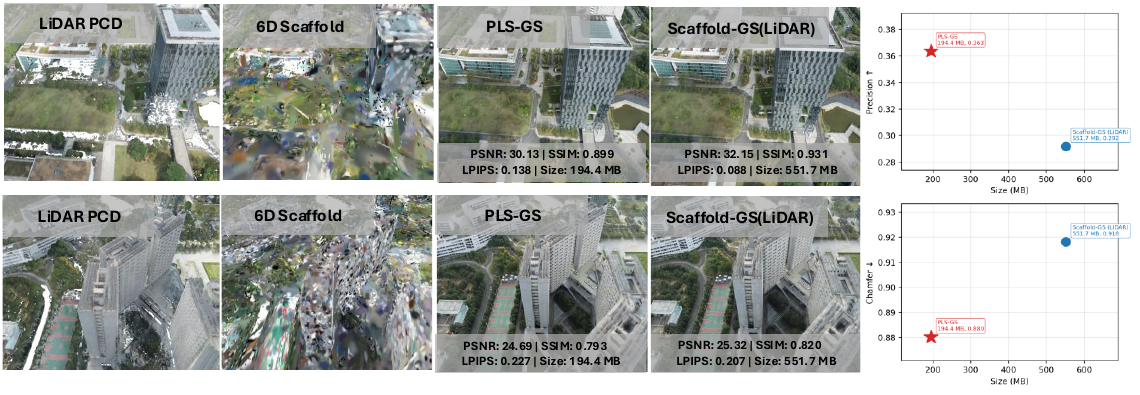}
    \caption{\textbf{\textit{PLS-GS} produces compact, surface-aligned scaffolds that place
    primitives closer to the measured surface.} Two held-out views of the College scene (rows). From
    the LiDAR point cloud and SfM input we build a continuous 6D HGMM scaffold (\emph{6D Scaffold})
    whose anchors are surface-aligned, in contrast to the unstructured anchor set of Scaffold-GS.
    \textit{PLS-GS} renders at $35\%$ of the model size of the LiDAR-initialized baseline. Right:
    it attains higher precision (\textcolor{red}{$\star$}) and lower Chamfer distance than that
    baseline (\textcolor{blue}{$\bullet$}) at a fraction of the size.}
    \label{fig:hero}
\end{figure*}

\begin{abstract}

While 3D Gaussian Splatting (3DGS) is typically optimized from images only, many robotics and mapping applications provide both camera and high-definition LiDAR data, combining rich visual appearance with accurate geometric structure. We introduce PLS-GS, a probabilistic LiDAR scaffold for Gaussian Splatting that exploits this complementary sensing setting to construct a compact 3DGS representation aligned to the measured surface. PLS-GS fits a Hierarchical Gaussian Mixture Model \emph{jointly} over position and color, yielding a continuous and adaptive scaffold that eliminates the need for discrete scene partitioning such as voxel grids or octrees. The fitted mixture then guides the renderable 3DGS Gaussians: each rendering Gaussian is associated with a GMM component and constrained in the local surface frame induced by that component's covariance. A Mahalanobis support loss bounds the deviation of the rendering Gaussians from the corresponding LiDAR-derived surface. Across five large outdoor scenes, PLS-GS places primitives markedly closer to the LiDAR measurements than both image- and LiDAR-initialized baselines, while employing substantially fewer primitives, yielding a considerably more compact model and faster rendering. At matched capacity, PLS-GS places primitives closer to the LiDAR measurements, with comparable rendering quality. Code is available at \url{https://github.com/k-harikumar/PLS-GS}.

\end{abstract}

\section{Introduction}
\label{sec:introduction}

3D Gaussian Splatting (3DGS)~\cite{kerbl2023gaussian} has rapidly become a leading representation for photorealistic novel-view synthesis, achieving high visual quality and real-time rendering with anisotropic Gaussian primitives optimized through differentiable rasterisation. While most 3DGS pipelines are driven by image supervision alone, many robotics and mapping platforms provide both cameras and high-definition LiDAR, where images supply rich appearance and LiDAR supplies accurate geometry. This complementarity creates an opportunity to construct 3DGS representations that retain high rendering quality while being more compact and more faithful to the scene geometry. We study 3DGS in exactly this setting.

Image-only 3DGS optimizes primitives to reproduce training views, which is effective for photometric quality but constrains geometry only indirectly: the learned primitives may be redundant and depart from the underlying surfaces~\cite{gauuscene}, and storage grows with scene scale~\cite{PapantonakisReduced3dgs}. Where measured geometry is available, it offers a direct way to ground the representation in the scene's true structure.

Two lines of work each capture part of what is needed. Scaffold-based methods~\cite{scaffoldgs, ren2024octree} improve compactness by decoding neural Gaussians from structured supports such as voxel grids or octrees, but these supports are discrete and seeded from image-derived structure rather than measured geometry. Separately, LiDAR-grounded methods~\cite{gssdf,ligs} improve alignment to measured geometry, but
their primitives remain free to drift under photometric optimization, retaining the unstructured
densification and per-primitive storage of vanilla 3DGS. Neither yields a scaffold that is both compact and grounded in measured geometry.

We propose PLS-GS, a probabilistic LiDAR scaffold for Gaussian Splatting that is continuous, adaptive to local scene structure, and grounded in measured geometry. PLS-GS fits a Hierarchical Gaussian Mixture Model \emph{jointly} over the position and color of the input point cloud, so that its components separate surfaces that are spatially adjacent but photometrically distinct. Unlike a discrete grid cell, each component carries a full covariance, representing not merely occupancy but the orientation and extent of the surface occupying it. Each component's covariance defines a local surface frame in which the neural Gaussians decoded from that anchor are expressed, and a Mahalanobis support loss bounds their deviation from it.

PLS-GS is a scaffolded Gaussian Splatting method that builds its supporting structure from measured LiDAR geometry instead of image-derived points. PLS-GS aligns its Gaussian primitives with the LiDAR measurements, resulting in a compact model that renders faster than baselines initialized from either images or LiDAR, with only a modest tradeoff in rendering PSNR. Concretely, our experiments validate the following claims:

\begin{itemize}
\item \textbf{Our probabilistic scaffold achieves substantially higher surface-alignment
precision than initialization from measured LiDAR geometry alone.} Seeding the Scaffold-GS baseline directly
from the LiDAR improves its alignment only marginally over image-derived points, whereas
PLS-GS improves it by $1.47\times$ (Sec.~\ref{sec:main_comparison}).

\item \textbf{Decoding primitives in the scaffold's local surface frame substantially improves
alignment to the measured geometry.} At identical anchor counts and identical anchor positions,
PLS-GS improves every geometric metric on every scene, at comparable rendering quality (Sec.~\ref{sec:matched}).

\item \textbf{Constrained probabilistic growth densifies under-reconstructed regions while keeping the representation compact.} PLS-GS grows its anchor count
by $1.1\times$ during training against Scaffold-GS's $2.7\times$, yielding a model
$2.5\times$ smaller that renders $1.7\times$ faster (Sec.~\ref{sec:main_comparison}).
\end{itemize}

\section{Related Works}
\label{sec:related_works}
\textbf{Gaussian Mixture Models for 3D Representation.}
Gaussian Mixture Models have a long history in point cloud modelling. Eckart et al.~\cite{eckarthgmm, Eckart_2018_ECCV} introduced hierarchical GMMs for multi-resolution representation, showing that recursive Expectation-Maximisation yields surface-aligned components whose covariance eigenvectors encode local surface geometry, a compact, continuous, and adaptive description of geometry in contrast to the discrete voxel grids and octrees commonly used to structure 3D data and reporting higher fidelity at smaller model size than NDT-based alternatives. To the best of our knowledge PLS-GS is the first method to repurpose this hierarchical GMM from a geometric representation into a rendering scaffold: its components serve as anchors, and their eigenvector frames define the surface-aligned frames in which neural Gaussians are decoded.

\textbf{Structured and Large-Scale Gaussian Splatting.}
3DGS~\cite{kerbl2023gaussian} enables real-time photorealistic rendering through differentiable
rasterisation of anisotropic Gaussian primitives, but produces large, redundant primitive sets.
Scaffold-GS~\cite{scaffoldgs} reduces this redundancy by decoding neural Gaussians from anchor
points through shared MLPs, and Octree-GS~\cite{ren2024octree} extends this with a
level-of-detail octree. Anchoring primitives to such discrete structures inherits the limitations
of discretisation: quantisation at cell boundaries, and a dependence on partitioning to scale. VastGaussian~\cite{vastgaussian},
CityGaussian~\cite{citygaussian} and Hierarchical-GS~\cite{hierarchicalgs} likewise address scale
by partitioning the scene into independently trained blocks. In all of these the supporting
structure is seeded and grown from photometric optimization rather than from scene geometry, so
the decoded primitives are not constrained to physical surfaces and image quality need not
reflect geometric accuracy~\cite{gauuscene}.

\textbf{Geometrically Accurate Gaussian Splatting.}
A separate line of work targets geometric accuracy through supervision during training. 2DGS~\cite{2dgs} and SuGaR~\cite{sugar} constrain primitives towards the underlying surface, and CityGaussianV2~\cite{citygaussianv2} attains strong accuracy on GauU-Scene V2 benchmark~\cite{gauuscene} with 2D Gaussian surfels and depth regression, at the cost of multi-GPU training, scene partitioning and mesh extraction. Closer to our setting, several methods supervise from measured LiDAR geometry. LI-GS~\cite{ligs} voxelises the cloud, fits a plane per voxel by RANSAC and estimates a mixture over in-plane position and intensity, so its components are planar by construction and inherit the discretisation of the underlying grid. ARSGaussian~\cite{Yao_2026} prunes primitives that drift beyond a distance threshold from the nearest measured point, GS-SDF~\cite{gssdf} regularises against a neural signed distance field, and Gaussian-LIC~\cite{gaussianlic} and GS-LIVO~\cite{gslivo} embed LiDAR within real-time SLAM. In each method, every Gaussian carries its own position, covariance, opacity and spherical-harmonic coefficients, so each primitive costs tens of floats to store.
\section{Preliminaries}
\label{sec:prelim}
\textbf{3D Gaussian Splatting.}
3DGS~\cite{kerbl2023gaussian} represents a scene as $N$ anisotropic Gaussian primitives, each with a centre $\boldsymbol{\mu}_i \in \mathbb{R}^3$, an opacity $\alpha_i$, view-dependent color, and a covariance factored into a rotation $\mathbf{R}_i$ and diagonal scaling $\mathbf{S}_i$ to preserve positive-definiteness during optimization:
\begin{equation}
\boldsymbol{\Sigma}_i = \mathbf{R}_i \mathbf{S}_i \mathbf{S}_i^{\top} \mathbf{R}_i^{\top}.
\label{eq:cov_decomp}
\end{equation}
Pixel colors are obtained by projecting the primitives to the image plane and compositing them front-to-back,
\begin{equation}
C(\mathbf{p}) = \sum_{i \in \mathcal{N}(\mathbf{p})} \mathbf{c}_i\, \tilde{\alpha}_i \prod_{j=1}^{i-1} (1 - \tilde{\alpha}_j),
\label{eq:alpha_compositing}
\end{equation}
where $\mathcal{N}(\mathbf{p})$ is the depth-sorted set of primitives covering $\mathbf{p}$ and $\tilde{\alpha}_i$ is $\alpha_i$ weighted by the projected 2D Gaussian at $\mathbf{p}$. All parameters are optimized against a photometric loss combining $L_1$ and structural-similarity terms:
\begin{equation}
\mathcal{L}_{\text{photo}} = (1 - \lambda) \mathcal{L}_1 + \lambda \mathcal{L}_{\text{SSIM}}.
\label{eq:photo_loss}
\end{equation}

\textbf{Scaffold-GS.}
Scaffold-GS~\cite{scaffoldgs} replaces the unstructured primitive set with a two-tier hierarchy of anchors and neural Gaussians. Each anchor $v$ at position $\mathbf{x}_v \in \mathbb{R}^3$ carries a learnable feature $\mathbf{f}_v$, a scaling vector $\mathbf{l}_v \in \mathbb{R}^6$, and $k$ offset vectors $\mathbf{O}_v \in \mathbb{R}^{k \times 3}$, where the first three components of $\mathbf{l}_v$ govern the offset extent and the remaining three the scale of the spawned Gaussians. Anchors are initialised by voxelising a sparse Structure-from-Motion point cloud $\mathcal{P}$ at resolution $\epsilon$,
\begin{equation}
\mathcal{V} = \left\{ \left\lfloor \tfrac{\mathbf{p}}{\epsilon} \right\rfloor \cdot \epsilon \;\middle|\; \mathbf{p} \in \mathcal{P} \right\},
\label{eq:scaffold_voxel}
\end{equation}
with duplicate centres removed. For each anchor within the view frustum, $k$ neural Gaussians are spawned at
\begin{equation}
\boldsymbol{\mu}_{v,i} = \mathbf{x}_v + \mathbf{O}_{v,i} \odot \mathbf{l}_v^{1:3}, \qquad i \in \{1, \ldots, k\},
\label{eq:scaffold_offset}
\end{equation}
their opacity, color, scale, and rotation decoded from $\mathbf{f}_v$ and the viewing parameters by shared MLPs and rasterised through Equations~\eqref{eq:cov_decomp}--\eqref{eq:alpha_compositing}. During training, Scaffold-GS densifies the anchor set at voxel centres where the averaged photometric gradient exceeds a threshold, so that the supporting structure, both its initial placement and its growth, is determined by a discrete voxel grid.

\textbf{Hierarchical Gaussian Mixture Models.}
A Gaussian mixture models a point set as a weighted sum of $J$ Gaussian densities,
\begin{equation}
p(\mathbf{y}) = \sum_{j=1}^{J} \pi_j \, \mathcal{N}(\mathbf{y} \mid \boldsymbol{\mu}_j, \boldsymbol{\Sigma}_j),
\qquad \textstyle\sum_j \pi_j = 1,
\label{eq:gmm}
\end{equation}
fitted by Expectation-Maximisation. Where points are sampled from a surface, a component's
covariance is anisotropic: its two largest eigenvalues span the local tangent plane and the
smallest aligns with the surface normal, so each component encodes the position, extent and
orientation of the surface it covers. Eckart et al.~\cite{eckarthgmm, Eckart_2018_ECCV} fit such a mixture recursively, refining each component over
the points assigned to it, so that a leaf $j$ at level $L$ carries the product of its ancestors'
weights,
\begin{equation}
p(\mathbf{y} \mid \Theta^{L}) = \sum_{j \in \mathcal{L}}
\Big( \textstyle\prod_{a \in \mathrm{path}(j)} \pi_a \Big)\,
\mathcal{N}(\mathbf{y} \mid \boldsymbol{\mu}_j, \boldsymbol{\Sigma}_j),
\label{eq:hgmm}
\end{equation}
and resolution follows local data density rather than a fixed cell size.
For more details about HGMM
construction, please see the supplementary material.

\section{PLS-GS}
\label{sec:method}

Given a colored point cloud, PLS-GS fits a Hierarchical Gaussian Mixture Model (HGMM) jointly over the spatial and color dimensions of the points, recursively refining a coarse mixture into progressively finer levels and taking the leaf components as anchors whose positions and orientations follow the local geometry of the scene. Unlike scaffold-based methods, which decode neural Gaussians from anchors on a discrete grid oriented in an arbitrary world frame, we decode each within the surface frame of its component and constrain it through a Mahalanobis support loss, so that both placement and rendering are governed by the measured geometry. A growth stage then adds anchors in regions of high photometric residual.

\subsection{Probabilistic Scaffold Construction}
\label{sec:scaffold}
Let the input be a colored point cloud $\mathcal{P} = \{ \mathbf{p}_i \}_{i=1}^{P}$, where each point $\mathbf{p}_i = [\mathbf{x}_i, \mathbf{c}_i] \in \mathbb{R}^6$ concatenates a spatial position $\mathbf{x}_i \in \mathbb{R}^3$ and a color $\mathbf{c}_i \in \mathbb{R}^3$. We standardise the six dimensions to zero mean and unit variance for fitting, so geometry and color contribute comparably.

We fit the HGMM level by level: from a single root, the points at each node are partitioned into $K=8$ children through Expectation-Maximisation, and each surviving component is in turn subdivided, producing a tree at increasing spatial and photometric resolution. A component is retained while the points assigned to it exceed a threshold, so the effective resolution adapts to local scene density rather than to a fixed grid. The leaf components form the scaffold.

Each leaf component $j$ yields a spatial mean $\boldsymbol{\mu}_j \in \mathbb{R}^3$, a spatial covariance $\boldsymbol{\Sigma}_j \in \mathbb{R}^{3\times3}$, and a color mean $\bar{\mathbf{c}}_j \in \mathbb{R}^3$. We instantiate one anchor per leaf, with position fixed at the mean,
\begin{equation}
\mathbf{x}_v = \boldsymbol{\mu}_j,
\label{eq:anchor_pos}
\end{equation}
and decompose the spatial covariance,
\begin{equation}
\boldsymbol{\Sigma}_j = \mathbf{U}_j \, \mathrm{diag}(\boldsymbol{\lambda}_j) \, \mathbf{U}_j^{\top},
\label{eq:eig}
\end{equation}
with the columns of $\mathbf{U}_j$ ordered by decreasing eigenvalue, so the first two span the local tangent plane and the third aligns with the surface normal. The eigenvectors $\mathbf{U}_j$ define the anchor's surface frame, the square-root eigenvalues $\sqrt{\boldsymbol{\lambda}_j}$ initialise its anisotropic scale, and the color mean $\bar{\mathbf{c}}_j$ initialises a per-anchor color (Section~\ref{sec:render}). These initial anchors' positions and surface frames are held fixed throughout training; only the per-anchor features, offsets, scales and colors, together with the shared MLP decoders, are optimized. New anchors are added during training through the probabilistic growth stage described in \cref{sec:growth}.

The joint treatment of geometry and color is essential: a mixture fitted to spatial coordinates alone cannot separate surfaces adjacent in space but distinct in appearance, such as a road marking against asphalt, whereas the color dimensions localise components on coherent physical surfaces, improving anchor placement and providing the color initialisation. Where the cloud is LiDAR, regions outside the sensor's coverage are filled with sparse Structure-from-Motion points so they can be rendered; these undergo the same construction and are not part of the geometric contribution.

\subsection{Surface-Aligned Rendering}
\label{sec:render}

\begin{figure}[H]
    \centering
    \includegraphics[width=\linewidth]{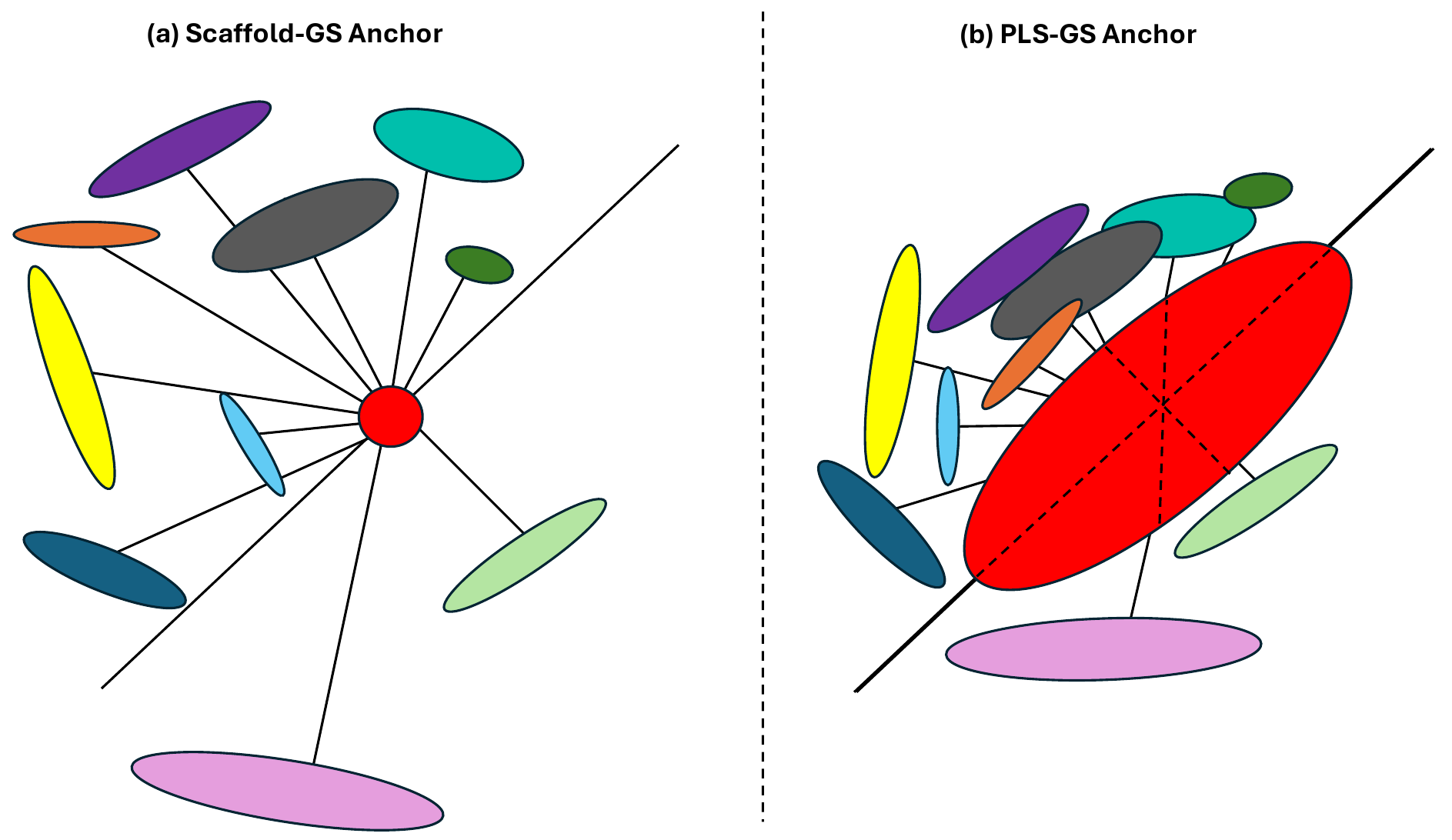}
    \caption{\textbf{Surface-aligned anchor formulation.} Scaffold-GS (left)
    anchors neural Gaussians at a point with world-axis offsets; \textit{PLS-GS}
    (right) anchors them in the component's eigenvector frame, aligning their
    spread with the local surface.}
    \label{fig:surface_align}
\end{figure}

Following Scaffold-GS, each anchor decodes $k$ neural Gaussians whose attributes are predicted from the anchor feature $\mathbf{f}_v$ and the viewing direction by shared MLPs. Whereas Scaffold-GS expresses each offset in world axes (Fig.~\ref{fig:surface_align}, left), we attach it to the anchor's surface frame (Fig.~\ref{fig:surface_align}, right). For learned offset $\mathbf{o}_{v,i} \in \mathbb{R}^3$ and offset extent $\mathbf{s}_v \in \mathbb{R}^3$, the neural Gaussian centre is
\begin{equation}
\boldsymbol{\mu}_{v,i} = \mathbf{x}_v + \mathbf{U}_v \, (\mathbf{o}_{v,i} \odot \mathbf{s}_v),
\label{eq:surface_offset}
\end{equation}
where $\mathbf{U}_v$ is the eigenvector frame of Equation~\eqref{eq:eig} and $\odot$ is the element-wise product. This aligns the two large axes of each anchor's spread with the local tangent plane, a geometric inductive bias that, together with the support loss (\cref{sec:support}), keeps decoded primitives near the measured surface while leaving them free in all three directions. Opacities, covariances, and colors are decoded as in Scaffold-GS, with the color mean $\bar{\mathbf{c}}_j$ as an additive base in logit space, so that at initialisation each anchor's rendered color equals that of its mixture component.

\subsection{Surface Support Loss}
\label{sec:support}
The surface frame orients the neural Gaussians but does not bound how far they drift from the anchor during optimization. We therefore add a support loss penalising each neural Gaussian's deviation by the Mahalanobis distance under the component's spatial covariance. For centre $\boldsymbol{\mu}_{v,i}$, let $\boldsymbol{\delta}_{v,i} = \boldsymbol{\mu}_{v,i} - \mathbf{x}_v$ be the displacement from the anchor, expressed in the local frame as $\tilde{\boldsymbol{\delta}}_{v,i} = \mathbf{U}_v^{\top} \boldsymbol{\delta}_{v,i}$. The loss averages the squared Mahalanobis distance over all rendered neural Gaussians,
\begin{equation}
\mathcal{L}_{\text{support}} = \frac{1}{M} \sum_{v,i} \sum_{d=1}^{3} \frac{ \tilde{\delta}_{v,i,d}^{\,2} }{ \lambda_{v,d} + \epsilon },
\label{eq:support}
\end{equation}

where $\lambda_{v,d}$ are the anchor's covariance eigenvalues, $M$ the number of rendered neural Gaussians, and $\epsilon$ a small constant. Because the penalty scales inversely with the eigenvalues, displacement is cheap along the high-variance tangent directions and costly along the low-variance normal, confining neural Gaussians to a thin neighbourhood of the surface while letting them spread across it. The training objective combines the photometric loss of Equation~\eqref{eq:photo_loss} with the support loss,
\begin{equation}
\mathcal{L} = \mathcal{L}_{\text{photo}} + \lambda_{\text{s}} \, \mathcal{L}_{\text{support}},
\label{eq:total_loss}
\end{equation}
where $\lambda_{\text{s}}$ balances rendering quality against geometric fidelity.

\subsection{Probabilistic Growth}
\label{sec:growth}

\begin{figure}[t]
    \centering
    \includegraphics[width=\linewidth]{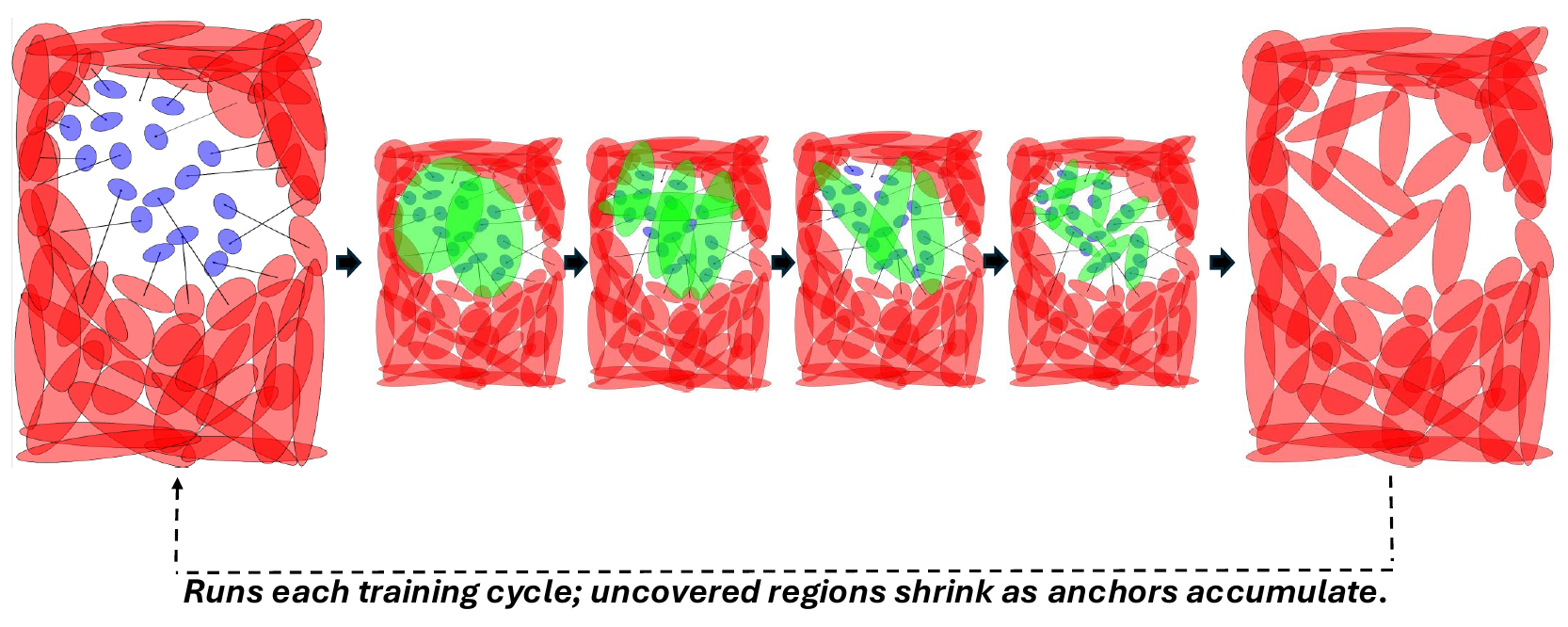}
    \caption{\textbf{HGMM-based anchor growth.} Neural Gaussians spawned in a region left uncovered
    by the scaffold accumulate high photometric residual (blue, left). A local HGMM is fitted to
    them and its leaf components are instantiated as new anchors (green), the mixture refining as
    the fit proceeds. The stage runs at each growth interval until training ends, so uncovered
    regions shrink progressively (right).}
    \label{fig:growth}
\end{figure}

The scaffold covers the measured surfaces, but some image regions, such as fine texture and high-frequency detail, are not fully explained by the initial anchors. We recover these through a growth stage using the same probabilistic construction.

During training we accumulate, per neural Gaussian, the average magnitude of its view-space positional gradient; those exceeding a threshold mark regions of persistent photometric residual. We pool their world-space positions and colors, fit a local six-dimensional HGMM as in Section~\ref{sec:scaffold}, and instantiate its leaf components as new anchors, with surface frames and scales from their covariances and deduplicated against the existing scaffold so covered regions are not repopulated. New anchors are marked as grown and may be removed if they fail to contribute, whereas measured-geometry anchors are retained throughout, preserving geometric fidelity.

Unlike the photometric densification of scaffold-based methods, which clones and splits anchors in proportion to the view-space gradient, this growth is bounded by construction: each event fits a mixture of fixed maximum depth, capping the anchors added regardless of residual magnitude, while the trigger is self-limiting, as the averaged gradient falls once a region is fitted. PLS-GS thus adds anchors only where the measured scaffold leaves genuine residual, central to the model's compactness.

\section{Experiments}
\label{sec:experiments}

\subsection{Dataset}
\label{sec:dataset}

We evaluate on GauU-Scene V2~\cite{gauuscene}, a large-scale outdoor benchmark providing dense survey-grade LiDAR alongside posed RGB images. To our knowledge it is the only public benchmark at this scale whose LiDAR is registered to the COLMAP frame recovered from the images of that scene. Hence measured LiDAR points can be used directly to initialize 3D Gaussian Splatting. 

We report five scenes, with approximate LiDAR point and camera counts: Campus ($80$M, $670$), Village ($27$M, $424$), College ($59$M, $1215$), Modern Building ($94$M, $713$), and Russian Building ($28$M, $563$). The five scenes span areas ranging from $0.8$ to $1.56\,\text{km}^2$. Images are downsampled to a longer side of $800$ pixels, and we hold out every eighth image as a test view, following prior work on this benchmark.

\subsection{Baselines}
\label{sec:baseline}
 We compare against Scaffold-GS~\cite{scaffoldgs} initialized from both SfM and LiDAR points.  Methods that supervise Gaussian Splatting with LiDAR~\cite{ligs,Yao_2026} would be the closest alternative comparison, but no implementation is publicly available at the time of writing. This comparison isolates the effect of replacing its discrete, photometrically grown scaffold with our continuous probabilistic scaffold and surface-aligned decoding. By also initializing baseline Scaffold-GS with LiDAR points, it isolates the contribution of the probabilistic 6D scaffold to alignment with the measured geometry. Further details, discussions and experiments regarding baseline selection are in the supplementary material.

\subsection{Implementation Details}
\label{sec:implementation}
Each cloud is fitted as in Section~\ref{sec:scaffold} by streaming GPU Expectation-Maximisation
with branching factor $K=8$, each level run to a relative log-likelihood tolerance of $10^{-4}$
capped at $20$ iterations, with covariance regularisation $10^{-4}$. Both the LiDAR and
Structure-from-Motion clouds are fitted to a leaf level of seven, discarding unsupported
components and those whose covariance eigenvalue aspect ratio exceeds $20$. The two mixtures are
unioned through a Mahalanobis gate on the joint position and color statistics at the $\chi^2$
radius for $0.99$ confidence, which removes Structure-from-Motion components already explained by
a LiDAR component; the LiDAR backbone is never removed. We build on the public Scaffold-GS
implementation~\cite{scaffoldgs}, replacing the world-frame offset with the surface-frame offset
(Eq.~\eqref{eq:surface_offset}), adding the support loss (Eq.~\eqref{eq:support}) and disabling
the per-image appearance embedding. Anchor positions and surface frames are fixed; all other
attributes and the decoder are optimised with Adam for $30{,}000$ iterations on a single
$24$\,GB GPU, with $\lambda_{\mathrm{s}}=10^{-3}$, $k=10$ and feature dimension $32$. Please see
the supplementary material for further details.

\subsection{Evaluation}
\label{sec:render_eval}
On the held-out test views we report PSNR, SSIM~\cite{1284395} and LPIPS~\cite{lpips}, with
anchor count, on-disk model size in MB, rendering speed in FPS, and training time to
characterise the capacity--quality trade-off.

\label{sec:geom_eval}
For geometric evaluation we follow CityGaussianV2~\cite{citygaussianv2}: a visibility-based crop
volume restricts the comparison to the well-observed region, and precision, recall and F-score
are computed against the LiDAR cloud at a per-scene distance threshold $\tau$, set to $1.5\times$
the mean nearest-neighbour distance of the cropped ground truth downsampled at $0.35\,$m, so the
tolerance reflects each scene's native sampling resolution. We lead with precision and Chamfer distance, which are normalised by the number of predicted
points and so cannot be improved by adding primitives; recall and F-score rise with primitive
count, and the methods compared here differ several-fold in capacity. Recall therefore measures
coverage rather than placement, and is not a meaningful basis for comparing representations of
different size; we report it, and the F-score it determines, only for consistency with prior work
on this benchmark. We therefore also report precision--recall curves, precision and F-score against the distance threshold, and the distribution of per-point prediction errors, together with a matched-capacity
comparison (Section~\ref{sec:matched}).

Geometry is extracted directly as the means of the decoded neural Gaussians, rather than through
the mesh extraction that surface-reconstruction methods apply before sampling, which for
volumetric anchors would measure the extraction algorithm as much as the representation. The same
extraction is applied to every method, so our scores are mutually comparable but not comparable to
mesh-sampled figures reported elsewhere. Further details regarding evaluation strategy are given in the supplementary material.
\section{Results}
\label{sec:results}

\begin{table*}[t]
\centering
\small
\setlength{\tabcolsep}{4pt}
\caption{\textbf{Main results on the five GauU-Scene V2 scenes.} Both baselines are
Scaffold-GS at its published configuration, differing only in the initialization cloud: SfM
points, or the survey LiDAR strided to the same anchor count. All methods use $d_f=32$, $k=10$
and 30K iterations at $800$-pixel resolution; PLS-GS additionally uses $\lambda_{\mathrm{s}} =
10^{-3}$. \emph{Init}/\emph{Fin} are anchor counts before and after training. Geometry is
evaluated on decoded neural-Gaussian means at the per-scene threshold $\tau$ under the protocol
of Section~\ref{sec:geom_eval}; Ch.\ is Chamfer distance in metres. Train is the optimization
loop only; Size is the model on disk in MB. Best per scene in \textbf{bold}.}
\label{tab:main}
\begin{tabular}{ll rr ccc cccc rrr}
\toprule
& & \multicolumn{2}{c}{Anchors} & \multicolumn{3}{c}{Rendering} & \multicolumn{4}{c}{Geometry (at $\tau$)} & \multicolumn{3}{c}{Efficiency} \\
\cmidrule(lr){3-4}\cmidrule(lr){5-7}\cmidrule(lr){8-11}\cmidrule(lr){12-14}
Scene & Method & Init & Fin & PSNR$\uparrow$ & SSIM$\uparrow$ & LPIPS$\downarrow$ & P$\uparrow$ & R$\uparrow$ & F1$\uparrow$ & Ch.$\downarrow$ & FPS$\uparrow$ & Train & Size$\downarrow$ \\
\midrule
\multirow{3}{*}{\shortstack[l]{\textbf{Campus}\\[1pt] \textcolor{gray}{\scriptsize $\tau=0.314$}}}
 & Scaffold-GS (SfM)   & 436K & 1370K & \textbf{27.71} & \textbf{0.863} & \textbf{0.175} & 0.195 & 0.102 & 0.134 & 1.190 & 123.9 & 22:33 & 432.9 \\
 & Scaffold-GS (LiDAR) & 436K & 1301K & 27.20 & 0.854 & 0.177 & 0.222 & \textbf{0.156} & 0.183 & 0.994 & 124.6 & 21:05 & 411.1 \\
 & PLS-GS              & 404K & \phantom{0}434K & 25.68 & 0.803 & 0.253 & \textbf{0.460} & 0.118 & \textbf{0.188} & \textbf{0.788} & \textbf{219.8} & \textbf{16:34} & \textbf{163.3} \\
\midrule
\multirow{3}{*}{\shortstack[l]{\textbf{Village}\\[1pt] \textcolor{gray}{\scriptsize $\tau=0.331$}}}
 & Scaffold-GS (SfM)   & 310K & 1052K & \textbf{27.77} & \textbf{0.875} & \textbf{0.143} & 0.388 & 0.273 & 0.320 & 0.673 & 130.5 & 20:39 & 332.3 \\
 & Scaffold-GS (LiDAR) & 311K & 1009K & 26.95 & 0.864 & 0.149 & 0.386 & \textbf{0.313} & \textbf{0.346} & 0.664 & 130.7 & 19:02 & 318.8 \\
 & PLS-GS              & 392K & \phantom{0}432K & 26.14 & 0.838 & 0.193 & \textbf{0.489} & 0.239 & 0.321 & \textbf{0.559} & \textbf{176.7} & \textbf{16:16} & \textbf{162.3} \\
\midrule
\multirow{3}{*}{\shortstack[l]{\textbf{College}\\[1pt] \textcolor{gray}{\scriptsize $\tau=0.333$}}}
 & Scaffold-GS (SfM)   & 855K & 1938K & \textbf{26.65} & \textbf{0.863} & \textbf{0.169} & 0.265 & 0.132 & 0.176 & 1.009 & 102.0 & 28:46 & 612.5 \\
 & Scaffold-GS (LiDAR) & 855K & 1746K & 25.87 & 0.853 & 0.172 & 0.292 & \textbf{0.178} & \textbf{0.221} & 0.918 & 109.7 & 25:45 & 551.7 \\
 & PLS-GS              & 487K & \phantom{0}517K & 25.11 & 0.812 & 0.230 & \textbf{0.363} & 0.094 & 0.150 & \textbf{0.880} & \textbf{211.6} & \textbf{17:27} & \textbf{194.4} \\
\midrule
\multirow{3}{*}{\shortstack[l]{\textbf{Modern}\\[1pt] \textcolor{gray}{\scriptsize $\tau=0.318$}}}
 & Scaffold-GS (SfM)   & 495K & 1302K & \textbf{27.30} & \textbf{0.852} & \textbf{0.157} & 0.358 & 0.179 & 0.239 & 0.811 & 120.6 & 32:00 & 411.3 \\
 & Scaffold-GS (LiDAR) & 496K & 1280K & 26.91 & 0.847 & 0.159 & 0.380 & \textbf{0.228} & \textbf{0.285} & 0.745 & 122.8 & 30:55 & 404.6 \\
 & PLS-GS              & 455K & \phantom{0}507K & 25.74 & 0.809 & 0.210 & \textbf{0.531} & 0.191 & 0.281 & \textbf{0.598} & \textbf{179.7} & \textbf{18:03} & \textbf{190.7} \\
\midrule
\multirow{3}{*}{\shortstack[l]{\textbf{Russian}\\[1pt] \textcolor{gray}{\scriptsize $\tau=0.338$}}}
 & Scaffold-GS (SfM)   & 464K & 1181K & \textbf{26.07} & \textbf{0.869} & \textbf{0.142} & 0.355 & 0.258 & 0.299 & 0.812 & \phantom{0}92.9 & 25:00 & 373.3 \\
 & Scaffold-GS (LiDAR) & 464K & 1202K & 25.48 & 0.857 & 0.146 & 0.369 & \textbf{0.352} & 0.360 & 0.779 & \phantom{0}84.3 & 29:02 & 379.9 \\
 & PLS-GS              & 340K & \phantom{0}409K & 24.56 & 0.822 & 0.197 & \textbf{0.578} & 0.313 & \textbf{0.406} & \textbf{0.688} & \textbf{164.3} & \textbf{20:25} & \textbf{153.8} \\
\midrule
\multirow{3}{*}{\textbf{Mean}}
 & Scaffold-GS (SfM)   & 512K & 1368K & \textbf{27.10} & \textbf{0.864} & \textbf{0.157} & 0.312 & 0.189 & 0.234 & 0.899 & 114.0 & 25:47 & 432.5 \\
 & Scaffold-GS (LiDAR) & 512K & 1308K & 26.48 & 0.855 & 0.161 & 0.330 & \textbf{0.245} & \textbf{0.279} & 0.820 & 114.4 & 25:09 & 413.2 \\
 & PLS-GS              & 416K & \phantom{0}460K & 25.45 & 0.817 & 0.217 & \textbf{0.484} & 0.191 & 0.269 & \textbf{0.703} & \textbf{190.4} & \textbf{17:45} & \textbf{172.9} \\
\bottomrule
\end{tabular}
\end{table*}

\begin{table*}[!t]
\centering
\small
\setlength{\tabcolsep}{4pt}
\caption{\textbf{Matched-capacity comparison across the five GauU-Scene V2 scenes.} Scaffold-GS
is initialized from the final anchor set of the corresponding trained PLS-GS model with growth
disabled, so both carry identical anchors at identical positions. Geometry is evaluated at the
per-scene threshold $\tau$ under the protocol of Section~\ref{sec:geom_eval}; Ch.\ is Chamfer
distance in metres. Size is given for PLS-GS as the per-anchor eigenvector frames plus the anchor
model, which is identical to the baseline's. Best per scene in \textbf{bold}.}
\label{tab:matched_capacity}
\begin{tabular}{ll rrrr cccc rrr}
\toprule
& & & \multicolumn{3}{c}{Rendering} & \multicolumn{4}{c}{Geometry (at $\tau$)} & \multicolumn{3}{c}{Efficiency} \\
\cmidrule(lr){4-6}\cmidrule(lr){7-10}\cmidrule(lr){11-13}
Scene & Method & Anchors & PSNR$\uparrow$ & SSIM$\uparrow$ & LPIPS$\downarrow$ & P$\uparrow$ & R$\uparrow$ & F1$\uparrow$ & Ch.$\downarrow$ & FPS$\uparrow$ & Train & Size$\downarrow$ \\
\midrule
\multirow{2}{*}{\shortstack[l]{\textbf{Campus}\\[1pt] \textcolor{gray}{\scriptsize $\tau=0.314$}}}
 & Scaffold-GS (static) & 434K & \textbf{25.89} & \textbf{0.814} & \textbf{0.234} & 0.251 & 0.082 & 0.124 & 1.324 & \textbf{311.6} & \textbf{10:36} & \textbf{137.2} \\
 & PLS-GS               & 434K & 25.68 & 0.803 & 0.253 & \textbf{0.460} & \textbf{0.118} & \textbf{0.188} & \textbf{0.788} & 219.8 & 16:34 & 26.1\,+\,137.2 \\
\midrule
\multirow{2}{*}{\shortstack[l]{\textbf{Village}\\[1pt] \textcolor{gray}{\scriptsize $\tau=0.331$}}}
 & Scaffold-GS (static) & 432K & \textbf{26.33} & 0.838 & \textbf{0.190} & 0.362 & 0.221 & 0.274 & 0.906 & \textbf{248.8} & \textbf{11:55} & \textbf{136.4} \\
 & PLS-GS               & 432K & 26.14 & 0.838 & 0.193 & \textbf{0.489} & \textbf{0.239} & \textbf{0.321} & \textbf{0.559} & 176.7 & 16:16 & 25.9\,+\,136.4 \\
\midrule
\multirow{2}{*}{\shortstack[l]{\textbf{College}\\[1pt] \textcolor{gray}{\scriptsize $\tau=0.333$}}}
 & Scaffold-GS (static) & 517K & 25.11 & \textbf{0.813} & \textbf{0.228} & 0.237 & 0.067 & 0.105 & 1.538 & \textbf{318.2} & \textbf{11:34} & \textbf{163.4} \\
 & PLS-GS               & 517K & 25.11 & 0.812 & 0.230 & \textbf{0.363} & \textbf{0.094} & \textbf{0.150} & \textbf{0.880} & 211.6 & 17:27 & 31.0\,+\,163.4 \\
\midrule
\multirow{2}{*}{\shortstack[l]{\textbf{Modern}\\[1pt] \textcolor{gray}{\scriptsize $\tau=0.318$}}}
 & Scaffold-GS (static) & 507K & \textbf{26.07} & \textbf{0.812} & \textbf{0.208} & 0.362 & 0.125 & 0.186 & 1.011 & \textbf{259.0} & \textbf{12:15} & \textbf{160.3} \\
 & PLS-GS               & 507K & 25.74 & 0.809 & 0.210 & \textbf{0.531} & \textbf{0.191} & \textbf{0.281} & \textbf{0.598} & 179.7 & 18:03 & 30.4\,+\,160.3 \\
\midrule
\multirow{2}{*}{\shortstack[l]{\textbf{Russian}\\[1pt] \textcolor{gray}{\scriptsize $\tau=0.338$}}}
 & Scaffold-GS (static) & 409K & \textbf{24.88} & \textbf{0.826} & \textbf{0.192} & 0.357 & 0.183 & 0.242 & 1.056 & \textbf{222.3} & \textbf{12:08} & \textbf{129.3} \\
 & PLS-GS               & 409K & 24.56 & 0.822 & 0.197 & \textbf{0.578} & \textbf{0.313} & \textbf{0.406} & \textbf{0.688} & 164.3 & 20:25 & 24.5\,+\,129.3 \\
\midrule
\multirow{2}{*}{\textbf{Mean}}
 & Scaffold-GS (static) & 460K & \textbf{25.66} & \textbf{0.821} & \textbf{0.210} & 0.314 & 0.136 & 0.186 & 1.167 & \textbf{272.0} & \textbf{11:42} & \textbf{145.3} \\
 & PLS-GS               & 460K & 25.45 & 0.817 & 0.217 & \textbf{0.484} & \textbf{0.191} & \textbf{0.269} & \textbf{0.703} & 190.4 & 17:45 & 27.6\,+\,145.3 \\
\bottomrule
\end{tabular}
\end{table*}

\begin{table*}[t]
\centering
\small
\setlength{\tabcolsep}{3.5pt}

\caption{\textbf{Scaffold construction, growth mechanism, and component ablation, College scene.}
Unmodified Scaffold-GS on three initializations, each with photometric growth disabled and
enabled; then the elements of PLS-GS added one at a time on the fused 6D scaffold, the
support-loss weight swept, and probabilistic growth applied at mixture leaf level four
(L4$^{\dagger}$, adopted) or five. The indented row repeats the adopted configuration without
anisotropic initialization. $\tau = 0.333$\,m; Ch.\ in metres; Size in MB. Best per column within each
block in \textbf{bold}. $^{\dagger}$adopted configuration.}
\label{tab:ablation}
\begin{tabular}{llrrrrrrrrrrrr}
\toprule
& & \multicolumn{2}{c}{Anchors} & \multicolumn{3}{c}{Rendering} & \multicolumn{4}{c}{Geometry (at $\tau$)} & \multicolumn{3}{c}{Efficiency} \\
\cmidrule(lr){3-4}\cmidrule(lr){5-7}\cmidrule(lr){8-11}\cmidrule(lr){12-14}
Configuration & Growth & Init & Final & PSNR$\uparrow$ & SSIM$\uparrow$ & LPIPS$\downarrow$ & P$\uparrow$ & R$\uparrow$ & F1$\uparrow$ & Ch.$\downarrow$ & FPS$\uparrow$ & Train & Size$\downarrow$ \\
\midrule
\multicolumn{14}{l}{\textbf{Scaffold-GS: photometric growth}} \\
\addlinespace[2pt]
SfM        & off & 855K & \phantom{0}855K & 25.98 & 0.837 & 0.200 & 0.195 & 0.086 & 0.119 & 1.243 & 202.9 & 22:06 & 270.2 \\
SfM        & on  & 855K & 1938K & \textbf{26.65} & \textbf{0.863} & \textbf{0.169} & 0.265 & 0.132 & 0.176 & 1.009 & 102.0 & 28:46 & 612.5 \\
LiDAR      & off & 855K & \phantom{0}855K & 24.98 & 0.817 & 0.215 & 0.240 & 0.110 & 0.151 & 1.204 & 210.7 & \textbf{17:19} & 270.2 \\
LiDAR      & on  & 855K & 1746K & 25.87 & 0.853 & 0.172 & \textbf{0.292} & \textbf{0.178} & \textbf{0.221} & \textbf{0.918} & 109.7 & 25:45 & 551.7 \\
6D HGMM means & off & 487K & \phantom{0}487K & 25.15 & 0.814 & 0.225 & 0.234 & 0.072 & 0.111 & 1.482 & \textbf{325.7} & 18:42 & \textbf{153.8} \\
6D HGMM means & on  & 487K & 1572K & 26.32 & 0.859 & 0.172 & 0.275 & 0.142 & 0.187 & 1.021 & 121.1 & 23:10 & 496.8 \\
\midrule
\multicolumn{14}{l}{\textbf{Probabilistic scaffold construction} (growth disabled)} \\
\addlinespace[2pt]
+ surface-frame rotation & off & 487K & \phantom{0}487K & 25.31 & 0.817 & 0.222 & 0.225 & 0.072 & 0.109 & 1.471 & \textbf{235.9} & \textbf{11:53} & 183.0 \\
+ color init             & off & 487K & \phantom{0}487K & \textbf{25.36} & \textbf{0.821} & \textbf{0.216} & \textbf{0.263} & \textbf{0.085} & \textbf{0.129} & \textbf{1.306} & 216.4 & 12:42 & 183.0 \\
+ anisotropic init       & off & 487K & \phantom{0}487K & 24.83 & 0.784 & 0.268 & 0.173 & 0.035 & 0.059 & 2.331 & 220.5 & 13:23 & 183.0 \\
\midrule
\multicolumn{14}{l}{\textbf{Mahalanobis support loss}} \\
\addlinespace[2pt]
$\lambda_{\mathrm{s}}=10^{-4}$ & off & 487K & \phantom{0}487K & 24.84 & 0.793 & 0.253 & 0.230 & 0.054 & 0.087 & 1.255 & 224.1 & 13:12 & 183.0 \\
$\lambda_{\mathrm{s}}=10^{-3}$$^{\dagger}$ & off & 487K & \phantom{0}487K & \textbf{24.98} & \textbf{0.807} & \textbf{0.238} & 0.392 & \textbf{0.093} & 0.150 & \textbf{0.791} & 242.7 & 12:15 & 183.0 \\
$\lambda_{\mathrm{s}}=10^{-2}$ & off & 487K & \phantom{0}487K & 24.79 & 0.794 & 0.259 & 0.439 & 0.092 & \textbf{0.152} & 0.798 & 238.0 & 11:59 & 183.0 \\
$\lambda_{\mathrm{s}}=10^{-1}$ & off & 487K & \phantom{0}487K & 23.85 & 0.735 & 0.336 & \textbf{0.547} & 0.057 & 0.103 & 0.899 & \textbf{250.3} & \textbf{11:17} & 183.0 \\
\midrule
\multicolumn{14}{l}{\textbf{PLS-GS: probabilistic growth}} \\
\addlinespace[2pt]
+ growth L4$^{\dagger}$  & L4 & 487K & \phantom{0}517K & \textbf{25.11} & \textbf{0.812} & \textbf{0.230} & \textbf{0.363} & \textbf{0.094} & \textbf{0.150} & 0.880 & 211.6 & \textbf{17:27} & 194.4 \\
\quad$-$ anisotropic init & L4 & 487K & \phantom{0}512K & 24.96 & 0.800 & 0.247 & 0.312 & 0.076 & 0.123 & 1.011 & \textbf{228.7} & 18:06 & 192.5 \\
+ growth L5              & L5 & 487K & \phantom{0}506K & 24.98 & 0.806 & 0.238 & 0.362 & 0.083 & 0.136 & \textbf{0.874} & 216.1 & 20:52 & \textbf{190.3} \\
\bottomrule
\end{tabular}
\end{table*}

\subsection{Main Comparison}
\label{sec:main_comparison}
Table~\ref{tab:main} reports rendering, geometry and efficiency across the five scenes against both
baselines. PLS-GS attains the highest precision and lowest Chamfer distance on every scene,
raising mean precision from $0.330$ to $0.484$ and lowering Chamfer distance from $0.820$ to
$0.703\,$m, with $2.8\times$ fewer anchors, $42\%$ of the storage, and $1.7\times$ the frame
rate. Initializing Scaffold-GS from the survey LiDAR rather than SfM points improves its alignment only
marginally, from $0.312$ to $0.330$ precision against $0.484$ for PLS-GS. The margin persists
when the baseline is handed our scaffold directly. Initialized from the 6D HGMM component means
and grown photometrically, it reaches $0.275$ precision and $1.021\,$m Chamfer distance on
College, against $0.363$ and $0.880\,$m for PLS-GS at a third of the anchors
(Table~\ref{tab:ablation}). The two methods reach these operating points differently. Both grow from photometric residual,
but Scaffold-GS instantiates new anchors on a discrete grid and grows $2.6$--$2.7\times$ during
training, while PLS-GS fits a local HGMM to the high-residual neural Gaussians and instantiates
its leaf components as anchors, growing $1.1\times$. The resulting difference in capacity is
what makes PLS-GS both smaller and faster to render, at $1.67\times$ the frame rate of
the baselines. It also explains recall and F-score, which scale with
capacity rather than with placement: the LiDAR baseline carries $2.8\times$ our anchors and
reaches $0.245$ recall against our $0.191$. Rendering quality follows the same pattern, PLS-GS
trailing by $1.03$--$1.65$\,dB PSNR compared to the baselines.

\begin{figure}[t]
    \centering
    \includegraphics[width=\linewidth]{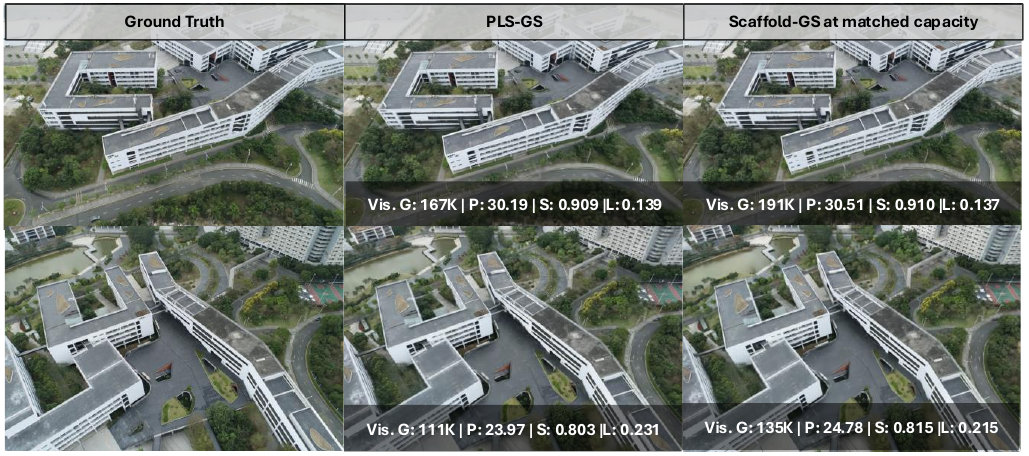}
    \caption{\textbf{Qualitative comparison on the College scene. } Two held-out views rendered by PLS-GS and by Scaffold-GS at matched capacity, against the ground truth. With identical anchor sets, the two render comparably, consistent with the 0.21 dB mean PSNR gap in Table 2. Badges report visible Gaussians (G), PSNR (P), SSIM (S) and LPIPS (L) for each view. Qualitative comparisons on all five scenes are provided in the supplementary material.}
    \label{fig:growth}
\end{figure}

\begin{figure}[t]
    \centering
    \includegraphics[width=\linewidth]{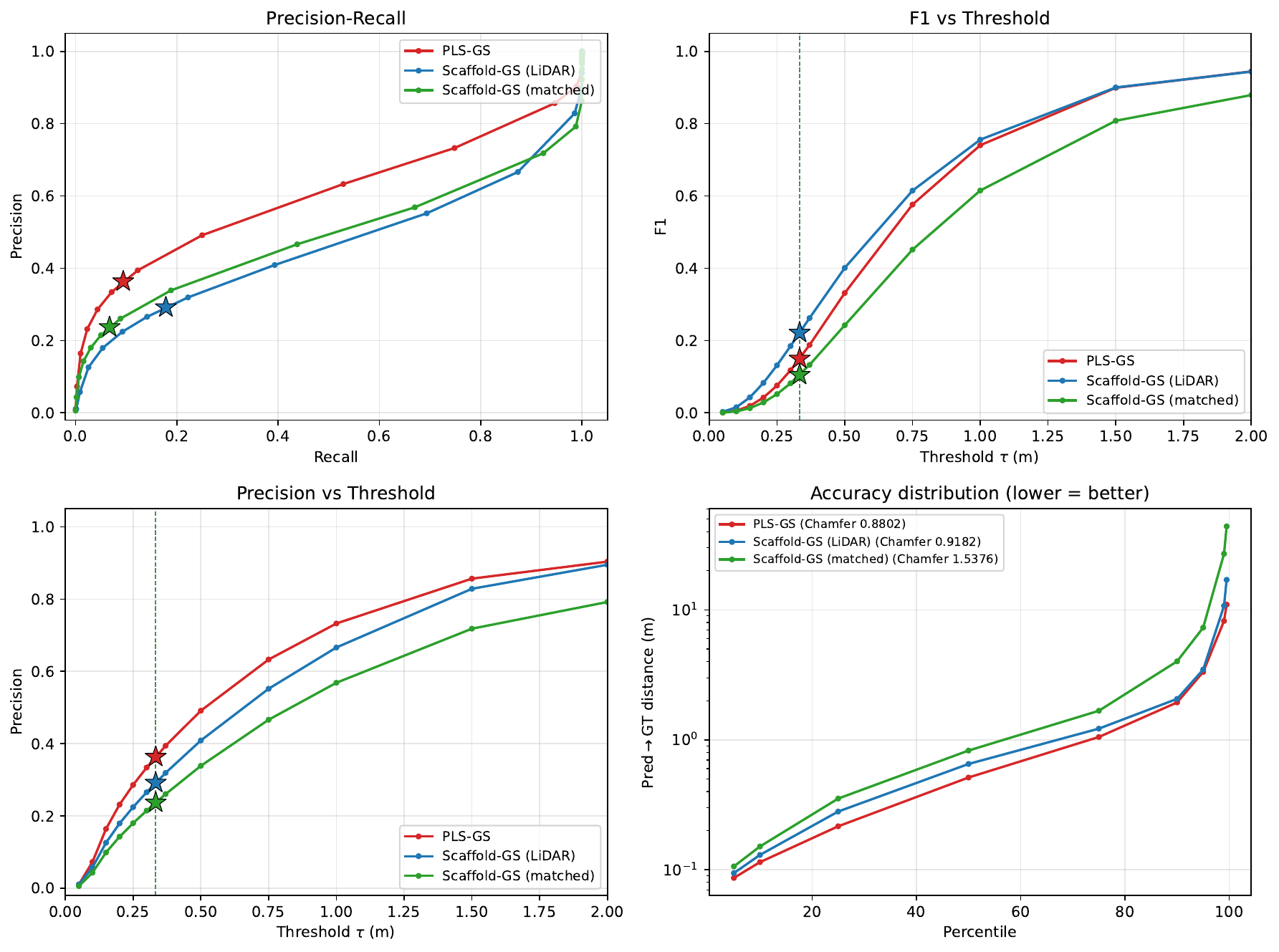}
    \caption{\textbf{Geometric alignment on the College scene.} Precision--recall, F1 and precision
    against the distance threshold $\tau$, and the distribution of per-point errors
    (Pred$\rightarrow$GT), for PLS-GS (\textcolor{red}{red}), the LiDAR-initialized baseline
    (\textcolor{blue}{blue}) and that baseline at matched capacity
    (\textcolor{OliveGreen}{green}). Stars mark the per-scene operating point $\tau$; threshold axes are truncated at
    $2$\,m.}
    \label{fig:geom_eval}
\end{figure}

\subsection{Matched-Capacity Experiments}
\label{sec:matched}
Table~\ref{tab:matched_capacity} initializes Scaffold-GS from the final anchor set of the trained
PLS-GS model with growth disabled, so both carry identical anchors at identical positions and
differ only in the decoding. PLS-GS then improves every geometric metric on every scene:
precision $1.54\times$, recall $1.40\times$, F-score $1.45\times$, Chamfer distance $-40\%$. The
recall and F-score advantages of Table~\ref{tab:main} therefore reflect capacity rather than
placement. The rendering difference collapses with them, from $1.03$ to $0.21\,$dB, with SSIM
within $0.004$ and LPIPS within $0.007$. The anchor points are identical in both cases; PLS-GS
additionally stores a per-anchor eigenvector frame, and the two terms are reported separately in
the table. Rendering speed is lower at matched capacity, $190$ against $272\,$FPS, due to the surface-aligned rendering formulation. 

\subsection{Component Ablation}
\label{sec:ablation}
Table~\ref{tab:ablation} ablates each stage in turn. Initializing from the survey LiDAR rather than SfM points raises precision $1.23\times$ but costs
$1.0\,$dB PSNR, as the aerial LiDAR does not cover everything the cameras observe; the fused 6D
scaffold partly recovers that rendering quality at $0.57\times$ the anchors. The surface frame alone does not improve placement ($0.225$ against $0.234$), since rotating an
unconstrained offset leaves its reachable set unchanged; under the support loss the same frame
raises precision $2.27\times$ and lowers Chamfer distance $2.95\times$ at unchanged anchor count. Anisotropic initialization outperforms
isotropic in the full method, where the loss makes the restriction purposeful. The
support loss acts as a regularizer on the photometric fit: up to $\lambda_{\mathrm{s}}=10^{-3}$
it improves PSNR, SSIM and LPIPS alongside precision, so geometric alignment is gained at no cost
in rendering. Probabilistic growth then adds $1.06\times$ capacity where photometric growth on
the same scaffold adds $3.23\times$, for a model $2.56\times$ smaller.

\subsection{Alignment Across Thresholds}
\label{sec:thresholds}
PLS-GS is more precise than the LiDAR-initialized baseline at every recall and threshold
(Figure~\ref{fig:geom_eval}), so the advantage is not an artefact of the chosen $\tau$, and the
per-point error distribution is tighter throughout rather than only in the tail. F1 favours the
baseline near $\tau$, as recall scales with its $3.4\times$ larger anchor count; at matched
capacity that advantage disappears. Qualitative comparisons on all five scenes, scaffold construction cost, and further support-loss and neural-Gaussian-count ablations are provided in the supplementary material.

\section{Limitations and Future Work}
\label{sec:limitations}
Despite the gains from probabilistic growth, rendering quality remains below Scaffold-GS with
unconstrained photometric growth. We plan to remedy this by exploiting the hierarchical nature of
the scaffold, refining the mixture and growing it more adaptively, and to investigate compression
strategies that maintain a compact representation under the added capacity. We also intend to investigate the effects of the Mahalanobis support loss against a standard rendered-depth loss in future work. A geometry-grounded scaffold that stays compact under growth is a natural
foundation for compact dynamic reconstruction, and we intend to extend the approach toward
geometrically faithful 4D Gaussian Splatting.

\section{Conclusions}
\label{sec:conclusion}
We introduced PLS-GS, a probabilistic LiDAR scaffold for Gaussian Splatting in which a
hierarchical Gaussian mixture is fitted jointly over position and color, and neural Gaussians
are decoded within the surface frame of each component. We show that a probabilistic scaffold preserves the geometry of the LiDAR point cloud more faithfully than initializing from that cloud directly, and that constrained probabilistic growth densifies under-reconstructed regions while maintaining this alignment and adding far less capacity than photometric growth, giving a model that is more compact and renders faster than the baselines.
\section*{Acknowledgements}
This work was supported by the Australian Research Council Research Hub in Intelligent Robotic Systems for Real-Time Asset Management (ARIAM) (IH210100030) and Emesent. The authors acknowledge the ongoing support from the Queensland University of Technology (QUT) Centre for Robotics.


{
    \small
    \bibliographystyle{ieeenat_fullname}
    \bibliography{main}

\begin{thebibliography}{20}
\providecommand{\natexlab}[1]{#1}
\providecommand{\url}[1]{\texttt{#1}}
\expandafter\ifx\csname urlstyle\endcsname\relax
  \providecommand{\doi}[1]{doi: #1}\else
  \providecommand{\doi}{doi: \begingroup \urlstyle{rm}\Url}\fi

\bibitem[Eckart et~al.(2016)Eckart, Kim, Troccoli, Kelly, and Kautz]{eckarthgmm}
Benjamin Eckart, Kihwan Kim, Alejandro Troccoli, Alonzo Kelly, and Jan Kautz.
\newblock {Accelerated Generative Models for 3D Point Cloud Data}.
\newblock In \emph{Proceedings of the IEEE/CVF Conference on Computer Vision and Pattern Recognition (CVPR)}, 2016.

\bibitem[Eckart et~al.(2018)Eckart, Kim, and Kautz]{Eckart_2018_ECCV}
Benjamin Eckart, Kihwan Kim, and Jan Kautz.
\newblock {HGMR: Hierarchical Gaussian Mixtures for Adaptive {3D} Registration}.
\newblock In \emph{Computer Vision – ECCV 2018: 15th European Conference, Munich, Germany, September 8-14, 2018, Proceedings, Part XV}, pages 730--746, Berlin, Heidelberg, 2018. Springer-Verlag.

\bibitem[Gu{\'e}don and Lepetit(2024)]{sugar}
Antoine Gu{\'e}don and Vincent Lepetit.
\newblock {SuGaR: Surface-Aligned Gaussian Splatting for Efficient 3D Mesh Reconstruction and High-Quality Mesh Rendering}.
\newblock In \emph{Proceedings of the IEEE/CVF Conference on Computer Vision and Pattern Recognition (CVPR)}, 2024.

\bibitem[Hong et~al.(2025)Hong, Zheng, Shen, Li, Zhang, Qin, and Shen]{gslivo}
Sheng Hong, Chunran Zheng, Yishu Shen, Changze Li, Fu Zhang, Tong Qin, and Shaojie Shen.
\newblock {GS-LIVO: Real-Time LiDAR, Inertial, and Visual Multisensor Fused Odometry With Gaussian Mapping}.
\newblock \emph{IEEE Transactions on Robotics}, 41:\penalty0 4253--4268, 2025.

\bibitem[Huang et~al.(2024)Huang, Yu, Chen, Geiger, and Gao]{2dgs}
Binbin Huang, Zehao Yu, Anpei Chen, Andreas Geiger, and Shenghua Gao.
\newblock {2D Gaussian Splatting for Geometrically Accurate Radiance Fields}.
\newblock In \emph{ACM SIGGRAPH Conference Proceedings}, 2024.

\bibitem[Jiang et~al.(2025)Jiang, Gao, Shao, Wang, Xiong, and Zhang]{ligs}
Changjian Jiang, Ruilan Gao, Kele Shao, Yue Wang, Rong Xiong, and Yu Zhang.
\newblock {LI-GS: Gaussian Splatting with LiDAR Incorporated for Accurate Large-Scale Reconstruction}.
\newblock \emph{IEEE Robotics and Automation Letters}, 10\penalty0 (2):\penalty0 1864--1871, 2025.

\bibitem[Kerbl et~al.(2023)Kerbl, Kopanas, Leimk{\"u}hler, and Drettakis]{kerbl2023gaussian}
Bernhard Kerbl, Georgios Kopanas, Thomas Leimk{\"u}hler, and George Drettakis.
\newblock {3D Gaussian Splatting for Real-Time Radiance Field Rendering}.
\newblock \emph{ACM Transactions on Graphics}, 42\penalty0 (4), 2023.

\bibitem[Kerbl et~al.(2024)Kerbl, Meuleman, Kopanas, Wimmer, Lanvin, and Drettakis]{hierarchicalgs}
Bernhard Kerbl, Andréas Meuleman, Georgios Kopanas, Michael Wimmer, Alexandre Lanvin, and George Drettakis.
\newblock {A Hierarchical 3D Gaussian Representation for Real-Time Rendering of Very Large Datasets}.
\newblock \emph{ACM Transactions on Graphics (SIGGRAPH Conference Proceedings)}, 43\penalty0 (4), 2024.

\bibitem[Lang et~al.(2025)Lang, Li, Wu, Zhao, Liu, Liu, Lv, and Zuo]{gaussianlic}
Xiaolei Lang, Laijian Li, Chenming Wu, Chen Zhao, Lina Liu, Yong Liu, Jiajun Lv, and Xingxing Zuo.
\newblock {Gaussian-LIC: Real-time photo-realistic SLAM with Gaussian splatting and LiDAR-inertial-camera fusion}.
\newblock In \emph{2025 IEEE International Conference on Robotics and Automation (ICRA)}, pages 8500--8507. IEEE, 2025.

\bibitem[Lin et~al.(2024)Lin, Li, Tang, Liu, Liu, Liu, Lu, Wu, Xu, Yan, and Yang]{vastgaussian}
Jiaqi Lin, Zhihao Li, Xiao Tang, Jianzhuang Liu, Shiyong Liu, Jiayue Liu, Yangdi Lu, Xiaofei Wu, Songcen Xu, Youliang Yan, and Wenming Yang.
\newblock {VastGaussian: Vast 3D Gaussians for Large Scene Reconstruction}.
\newblock In \emph{Proceedings of the IEEE/CVF Conference on Computer Vision and Pattern Recognition (CVPR)}, 2024.

\bibitem[Liu et~al.(2025{\natexlab{a}})Liu, Wan, Wang, Zheng, Lin, and Zhang]{gssdf}
Jianheng Liu, Yunfei Wan, Bowen Wang, Chunran Zheng, Jiarong Lin, and Fu Zhang.
\newblock {GS-SDF: LiDAR-Augmented Gaussian Splatting and Neural SDF for Geometrically Consistent Rendering and Reconstruction}.
\newblock In \emph{2025 IEEE/RSJ International Conference on Intelligent Robots and Systems (IROS)}, pages 19391--19398. IEEE, 2025{\natexlab{a}}.

\bibitem[Liu et~al.(2025{\natexlab{b}})Liu, Luo, Fan, Wang, Peng, and Zhang]{citygaussian}
Yang Liu, Chuanchen Luo, Lue Fan, Naiyan Wang, Junran Peng, and Zhaoxiang Zhang.
\newblock {CityGaussian: Real-time High-quality Large-Scale Scene Rendering with Gaussians}.
\newblock In \emph{European Conference on Computer Vision}, pages 265--282. Springer, 2025{\natexlab{b}}.

\bibitem[Liu et~al.(2025{\natexlab{c}})Liu, Luo, Mao, Peng, and Zhang]{citygaussianv2}
Yang Liu, Chuanchen Luo, Zhongkai Mao, Junran Peng, and Zhaoxiang Zhang.
\newblock {CityGaussianV2: Efficient and Geometrically Accurate Reconstruction for Large-Scale Scenes}.
\newblock In \emph{The Thirteenth International Conference on Learning Representations}, 2025{\natexlab{c}}.

\bibitem[Lu et~al.(2024)Lu, Yu, Xu, Xiangli, Wang, Lin, and Dai]{scaffoldgs}
Tao Lu, Mulin Yu, Linning Xu, Yuanbo Xiangli, Limin Wang, Dahua Lin, and Bo Dai.
\newblock {Scaffold-{GS}: Structured 3D Gaussians for View-Adaptive Rendering}.
\newblock In \emph{Proceedings of the IEEE/CVF Conference on Computer Vision and Pattern Recognition (CVPR)}, 2024.

\bibitem[Papantonakis et~al.(2024)Papantonakis, Kopanas, Kerbl, Lanvin, and Drettakis]{PapantonakisReduced3dgs}
Panagiotis Papantonakis, Georgios Kopanas, Bernhard Kerbl, Alexandre Lanvin, and George Drettakis.
\newblock {Reducing the Memory Footprint of 3D Gaussian Splatting}.
\newblock \emph{Proceedings of the ACM on Computer Graphics and Interactive Techniques}, 7\penalty0 (1), 2024.

\bibitem[Ren et~al.(2025)Ren, Jiang, Lu, Yu, Xu, Ni, and Dai]{ren2024octree}
Kerui Ren, Lihan Jiang, Tao Lu, Mulin Yu, Linning Xu, Zhangkai Ni, and Bo Dai.
\newblock {Octree-GS: Towards Consistent Real-time Rendering with LOD-Structured 3D Gaussians}.
\newblock \emph{IEEE Transactions on Pattern Analysis and Machine Intelligence}, pages 1--15, 2025.

\bibitem[Wang et~al.(2004)Wang, Bovik, Sheikh, and Simoncelli]{1284395}
Zhou Wang, A.C. Bovik, H.R. Sheikh, and E.P. Simoncelli.
\newblock {Image quality assessment: from error visibility to structural similarity}.
\newblock \emph{IEEE Transactions on Image Processing}, 13\penalty0 (4):\penalty0 600--612, 2004.

\bibitem[Xiong et~al.(2024)Xiong, Zheng, Liu, and Li]{gauuscene}
Butian Xiong, Nanjun Zheng, Junhua Liu, and Zhen Li.
\newblock {GauU-Scene V2: Assessing the Reliability of Image-Based Metrics with Expansive Lidar Image Dataset Using 3DGS and NeRF}.
\newblock \emph{arXiv preprint arXiv:2404.04880}, 2024.

\bibitem[Yao et~al.(2026)Yao, Zhang, Zhang, Gao, Peng, Li, Wang, and Wang]{Yao_2026}
Yiling Yao, Bing Zhang, Wenjuan Zhang, Lianru Gao, Dailiang Peng, Bocheng Li, Yaning Wang, and Bowen Wang.
\newblock {ARSGaussian: 3D Gaussian Splatting with LiDAR for Aerial Remote Sensing Novel View Synthesis}.
\newblock \emph{ISPRS Journal of Photogrammetry and Remote Sensing}, 231:\penalty0 288--306, 2026.

\bibitem[Zhang et~al.(2018)Zhang, Isola, Efros, Shechtman, and Wang]{lpips}
Richard Zhang, Phillip Isola, Alexei~A. Efros, Eli Shechtman, and Oliver Wang.
\newblock {The Unreasonable Effectiveness of Deep Features as a Perceptual Metric}.
\newblock In \emph{Proceedings of the IEEE/CVF Conference on Computer Vision and Pattern Recognition (CVPR)}, 2018.

\end{thebibliography}
}

\clearpage
\onecolumn
\setcounter{section}{0}\setcounter{table}{0}\setcounter{figure}{0}\setcounter{equation}{0}
\renewcommand{\thesection}{S-\Roman{section}}
\renewcommand{\thetable}{S\arabic{table}}
\renewcommand{\thefigure}{S\arabic{figure}}
\renewcommand{\theequation}{S\arabic{equation}}
\renewcommand{\theHsection}{supp.\arabic{section}}
\renewcommand{\theHtable}{supp.\arabic{table}}
\renewcommand{\theHfigure}{supp.\arabic{figure}}
\renewcommand{\theHequation}{supp.\arabic{equation}}
\begin{center}
{\Large\bf \textit{PLS-GS}: Probabilistic Lidar Scaffolds for\\Compact, Surface-Aligned Gaussian Splatting\\[0.3em]
{\large Supplementary Material}}
\end{center}
\vspace{1em}
\suppressfloats[t]  

\section{Implementation and Evaluation Details}
\label{sec:supp_impl}
 
This section details the baseline configuration, scaffold construction, training schedule, and
evaluation protocol summarized in the main paper.
 
\subsection{Baseline Configuration}
\label{sec:supp_baseline}

Both baselines are the public Scaffold-GS implementation, unmodified. We use adaptive
voxelization ($\texttt{voxel\_size}=-1$), which selects the initialization voxel from the
nearest-neighbour statistics of the input cloud, as Scaffold-GS does for large-scale scenes, and
no point stride ($\texttt{ratio}=1$). Both baselines use the same resolution, iteration count,
feature dimension and $k$ as PLS-GS.

The LiDAR-initialized baseline is seeded from the survey cloud strided uniformly to the anchor
count of the SfM-initialized arm, so that the two begin from equivalent capacity and differ only
in the source of their supporting structure. Strides are Campus $183$, Village $86$, College
$69$, Modern $190$ and Russian $61$.

The matched-capacity variant of the main paper initializes Scaffold-GS from the final anchor set
of the corresponding trained PLS-GS model with $\texttt{update\_until}=0$, which disables both
anchor growth and opacity pruning, so the final anchor count equals the initial.
 
\subsection{Scaffold Construction}
\label{sec:supp_scaffold}
 
\begin{figure*}[t]
  \centering
  \includegraphics[width=0.75\linewidth]{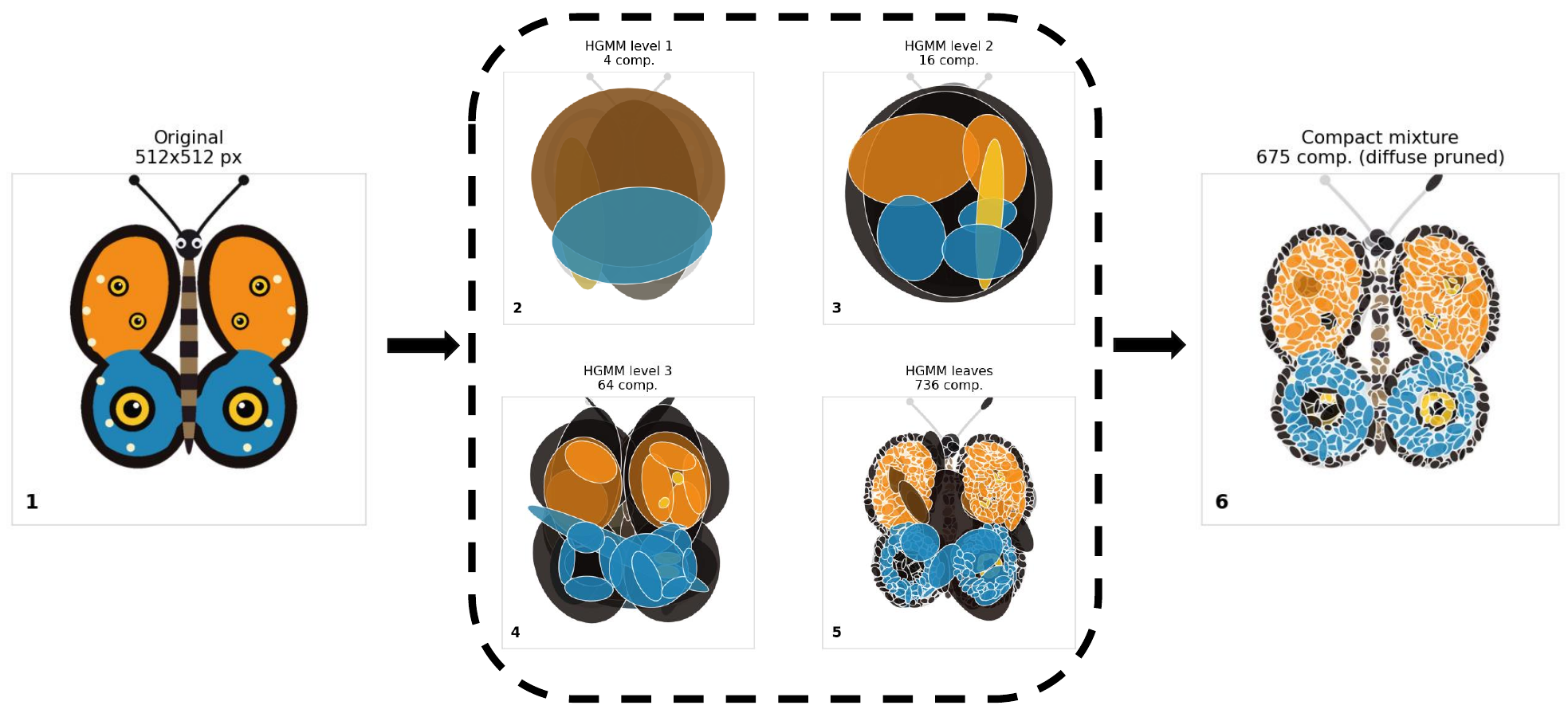}
  \caption{\textbf{Conceptual illustration of a hierarchical Gaussian mixture}, shown on a
  2D image for intuition: a coarse-to-fine mixture (1--5) is pruned to a compact set (6).
  \emph{(Schematic only --- the branching factor, level counts, and component numbers shown are
  illustrative and do not reflect the HGMM configuration, dimensionality (6D), or fitting
  used in PLS-GS.)}}
  \label{fig:hgmm}
\end{figure*}
 
\paragraph{Input preparation.}
The LiDAR cloud is split into position and color arrays, with an xyz-only copy retained as the
geometric reference. The SfM cloud is filtered for floaters by statistical outlier removal
($\text{nb}=20$, $\text{std}=2.0$) before fitting. As GauU-Scene V2 supplies the LiDAR already
registered to the COLMAP frame, both clouds are fitted in that frame, and the resulting scaffold
covers the scene and can be used directly to initialize Gaussian Splatting.
 
\paragraph{Hierarchical mixture fitting.}
The hierarchical Gaussian mixture is fitted on the GPU by streaming Expectation-Maximization over
the six-dimensional points (position and color), standardized to zero mean and unit variance per
dimension before fitting and returned to the input frame afterward. Each node is partitioned into
$K=8$ children, and each level is iterated to a relative log-likelihood tolerance of $10^{-4}$
subject to a cap of $20$ iterations, with a covariance regularization of $10^{-4}$ added to
maintain full-rank components. Both clouds are fitted to a leaf level of seven, with a minimum
component support of $32$ points for the LiDAR mixture and $1$ for the SfM mixture. Components with fewer points than the respective minimum support are discarded, as are components whose spatial-covariance eigenvalue aspect ratio exceeds 20, removing highly elongated components that would otherwise produce large, poorly localized Gaussians.
 
\paragraph{Redundancy gate.}
The scaffold unions the LiDAR mixture and the SfM mixture. Because the two overlap wherever the
LiDAR is well observed, each SfM component is tested for redundancy against the LiDAR mixture
before inclusion. Treating the LiDAR mixture as a reference, an SfM component is compared to its
nearest LiDAR components through a Mahalanobis gate on the joint six-dimensional
position--color statistics: it is removed if it falls within the gate of any LiDAR component and
retained otherwise. The gate radius is the $\chi^2$ value at $0.99$ confidence with six degrees
of freedom. The LiDAR backbone is never removed, so SfM components survive only where no LiDAR
component explains their location or color, populating regions outside LiDAR coverage without a
hand-tuned spatial crop.
 
\paragraph{Construction cost.}
Table~\ref{tab:construction_cost} reports the cost of building the scaffold. Fitting time scales roughly
linearly with input point count, while peak memory is dominated by the chunk size and the number
of leaf nodes rather than by the size of the cloud, and is therefore close to constant across
scenes. Construction is a one-time preprocessing cost, incurred once per scene before any
training run, and the two mixtures are fitted sequentially, so the binding hardware requirement
is the larger of the two peaks rather than their sum.

\begin{table*}[t]
\centering
\small
\setlength{\tabcolsep}{5pt}
\caption{\textbf{Scaffold construction and training cost.} PLS-GS fits the 6D HGMM scaffold once
per scene as a preprocessing step; Scaffold-GS voxelizes its initialization cloud inline and has
no equivalent stage. \emph{HGMM fit} is the LiDAR and SfM mixtures combined, with peak memory
measured as the CuPy pool high-water mark; \emph{Training} is the 30K-iteration optimization loop,
with peak memory from \texttt{torch.cuda.max\_memory\_reserved}. Both on a single $24$\,GB GPU.
The two stages run sequentially, so the \emph{Total} peak is the larger of the two stage peaks,
not their sum. Best total per scene in \textbf{bold}.}
\label{tab:construction_cost}
\begin{tabular}{llr rr rr rr}
\toprule
& & & \multicolumn{2}{c}{HGMM fit} & \multicolumn{2}{c}{Training} & \multicolumn{2}{c}{Total} \\
\cmidrule(lr){4-5}\cmidrule(lr){6-7}\cmidrule(lr){8-9}
Scene & Method & LiDAR pts & Time & Peak (GB) & Time & Peak (GB) & Time & Peak (GB) \\
\midrule
\multirow{3}{*}{\textbf{Campus}} & Scaffold-GS (SfM)   & \multirow{3}{*}{79.8M} & -- & -- & 22:33 & 10.09 & 22:33 & 10.09 \\
 & Scaffold-GS (LiDAR) & & -- & -- & 21:05 & 10.42 & \textbf{21:05} & 10.42 \\
 & PLS-GS              & & 33:20 & 10.05 & 16:34 & \phantom{0}7.61 & 49:54 & \textbf{10.05} \\
\midrule
\multirow{3}{*}{\textbf{Village}} & Scaffold-GS (SfM)   & \multirow{3}{*}{26.8M} & -- & -- & 20:39 & \phantom{0}7.86 & 20:39 & \phantom{0}7.86 \\
 & Scaffold-GS (LiDAR) & & -- & -- & 19:02 & \phantom{0}7.21 & \textbf{19:02} & \phantom{0}\textbf{7.21} \\
 & PLS-GS              & & 14:58 & 13.33 & 16:16 & \phantom{0}6.38 & 31:14 & 13.33 \\
\midrule
\multirow{3}{*}{\textbf{College}} & Scaffold-GS (SfM)   & \multirow{3}{*}{59.0M} & -- & -- & 28:46 & 14.75 & 28:46 & 14.75 \\
 & Scaffold-GS (LiDAR) & & -- & -- & 25:45 & 14.46 & \textbf{25:45} & 14.46 \\
 & PLS-GS              & & 26:19 & 10.78 & 17:27 & 11.58 & 43:46 & \textbf{11.58} \\
\midrule
\multirow{3}{*}{\textbf{Modern}} & Scaffold-GS (SfM)   & \multirow{3}{*}{94.2M} & -- & -- & 32:00 & \phantom{0}9.91 & 32:00 & \phantom{0}\textbf{9.91} \\
 & Scaffold-GS (LiDAR) & & -- & -- & 30:55 & 10.22 & \textbf{30:55} & 10.22 \\
 & PLS-GS              & & 39:41 & 11.24 & 18:03 & \phantom{0}8.12 & 57:44 & 11.24 \\
\midrule
\multirow{3}{*}{\textbf{Russian}} & Scaffold-GS (SfM)   & \multirow{3}{*}{28.3M} & -- & -- & 25:00 & 10.13 & \textbf{25:00} & \textbf{10.13} \\
 & Scaffold-GS (LiDAR) & & -- & -- & 29:02 & 10.71 & 29:02 & 10.71 \\
 & PLS-GS              & & 16:03 & 10.33 & 20:25 & \phantom{0}7.07 & 36:28 & 10.33 \\
\midrule
\multirow{3}{*}{\textbf{Mean}} & Scaffold-GS (SfM)   & \multirow{3}{*}{57.6M} & -- & -- & 25:47 & 10.55 & 25:47 & \textbf{10.55} \\
 & Scaffold-GS (LiDAR) & & -- & -- & 25:09 & 10.60 & \textbf{25:09} & 10.60 \\
 & PLS-GS              & & 26:04 & 11.15 & 17:45 & \phantom{0}8.15 & 43:49 & 11.31 \\
\bottomrule
\end{tabular}
\end{table*}

\subsection{Training}
\label{sec:supp_training}
 
We build on the public Scaffold-GS implementation, retaining its decoder and rasterization
pipeline, replacing the world-frame offset with the surface-frame offset, and adding the
Mahalanobis support loss and the growth stage. Each anchor carries a feature of dimension $32$
and decodes $k=10$ neural Gaussians. The initial anchors' positions and surface frames are held fixed
throughout training; only the per-anchor features, offsets,
scales and colors, together with the shared MLP decoders,
are optimized with Adam for $30{,}000$ iterations on a single
$24$\,GB GPU, with support-loss weight $\lambda_{\text{s}}=10^{-3}$. The per-image
appearance embedding of Scaffold-GS is disabled to isolate the geometric scaffold. The growth
stage accumulates view-space positional gradients from iteration $500$, and from iteration
$1{,}500$ to $15{,}000$ fits a local six-dimensional HGMM to high-residual neural Gaussians every
$100$ iterations, instantiating their leaf components as new anchors; anchors introduced by
growth may be pruned if they fail to contribute, whereas measured-geometry anchors are retained
throughout.
 
\subsection{Geometric Evaluation Protocol}
\label{sec:supp_eval}
 
We evaluate geometry against the ground-truth LiDAR cloud following the crop-and-threshold
protocol of CityGaussianV2. A visibility-based crop volume restricts the comparison to the
well-observed region: ground-truth points are rasterized across the training views, those whose
accumulated visibility exceeds a threshold are retained, and an alpha shape is fitted to the
result and exported as a selection volume. As camera counts vary across scenes, the visibility
threshold is set per scene to the mean per-point view count rather than to a fixed absolute
value, keeping the retained region the well-observed core of each scene.
 
The distance threshold $\tau$ is derived per scene from ground-truth statistics: the cropped
ground truth is downsampled at a $0.35\,\text{m}$ voxel, and $\tau$ is set to $1.5\times$ the
mean nearest-neighbour distance of the downsampled cloud, so the tolerance reflects each scene's
native sampling resolution rather than a hand-picked value. The resulting thresholds are Campus
$0.314$, Village $0.331$, College $0.333$, Modern $0.318$, and Russian $0.338\,\text{m}$, all
within the $0.3$--$0.6\,\text{m}$ range reported by CityGaussianV2. All distances are converted
to metres using a per-scene similarity-transform scale factor that maps COLMAP units to metres,
estimated once per scene and held fixed across all experiments (Campus $92.46$, Village $67.78$,
College $116.19$, Modern $70.59$, Russian $79.24$). Geometry is extracted directly as the means
of the decoded neural Gaussians, with no intervening mesh extraction, and every method including
the baselines is evaluated under this same extraction so that scores are mutually comparable.
Precision, recall, and F-score are computed at $\tau$, together with the bidirectional Chamfer
distance.
 
\section{Additional Results}
\label{sec:supp_results}
 
\subsection{Qualitative Rendering Comparison}
\label{sec:supp_qual}
 
\begin{figure*}[t]
\centering
\includegraphics[width=\textwidth]{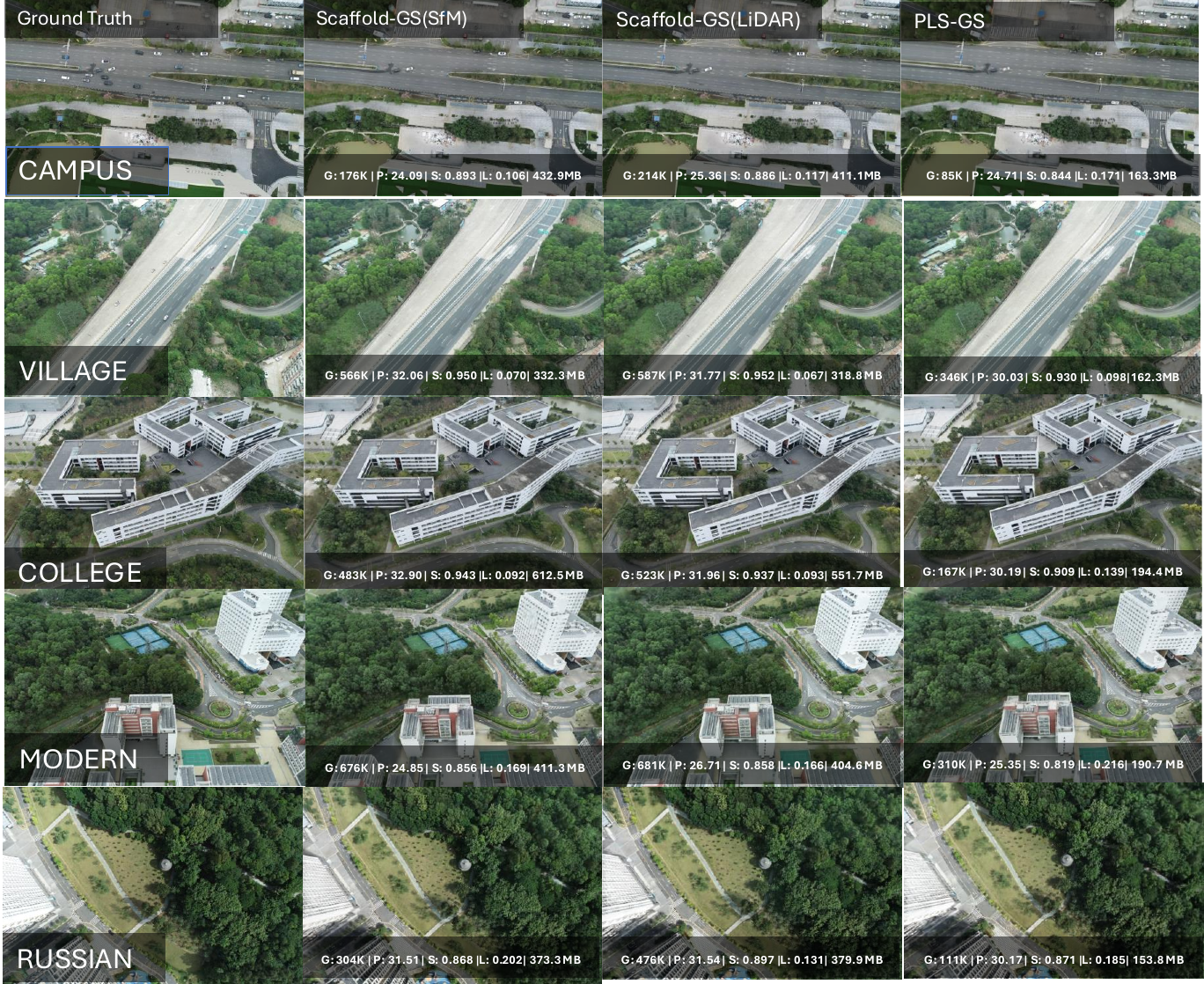}
\caption{\textbf{Qualitative rendering comparison across the five GauU-Scene V2 scenes.} One
held-out view per scene: ground truth, Scaffold-GS initialized from SfM points, Scaffold-GS
initialized from the survey LiDAR, and PLS-GS. Badges report the number of Gaussians visible in
that view (G), PSNR (P), SSIM (S), LPIPS (L), and the on-disk model size.}
\label{fig:supp_qual}
\end{figure*}

Figure~\ref{fig:supp_qual} compares held-out renderings across all five scenes. PLS-GS rasterizes fewer Gaussians than the baselines in every view, and fewer than half as many in four of the five. This is reflected in the rendering-speed difference shown in Table~1 of the main paper: fewer primitives reach the rasterizer per frame, so fewer are sorted and blended.
 
\subsection{Support-Loss Weight After Growth}
\label{sec:supp_lambda}
 
The main paper sweeps $\lambda_{\text{s}}$ with growth disabled, isolating the loss at fixed
capacity. Table~\ref{tab:lambda_sweep} repeats the sweep in the full method with growth enabled,
that is, at the configuration actually deployed.
 
\begin{figure}[t]
\centering
\includegraphics[width=\linewidth]{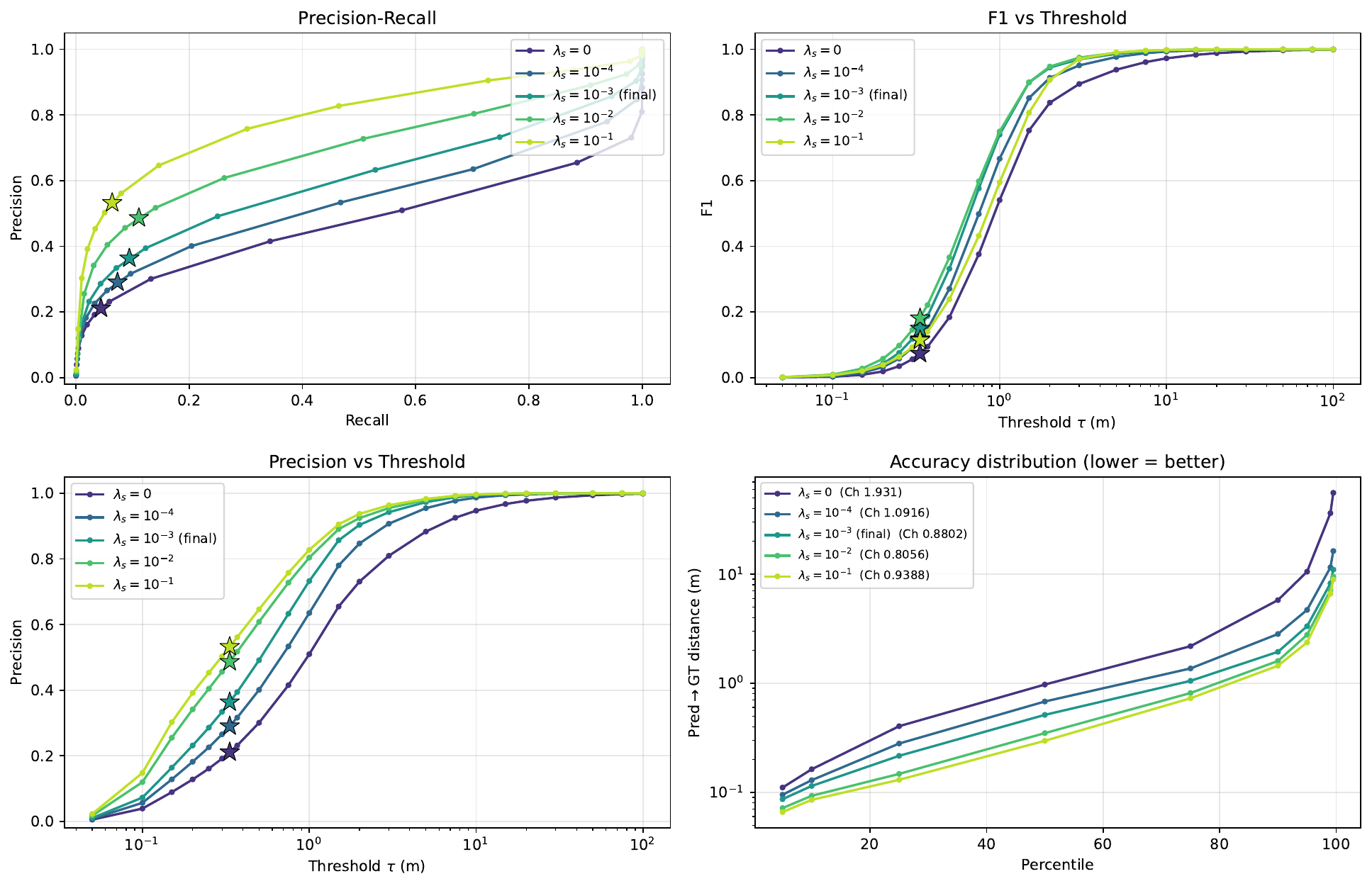}
\caption{\textbf{Geometric effect of the support-loss weight} $\lambda_{\text{s}}$ on the College
scene, across the same sweep as Table~\ref{tab:lambda_sweep}. Curves are colored from
$\lambda_{\text{s}}=0$ (dark) to $\lambda_{\text{s}}=10^{-1}$ (light); stars mark the per-scene
operating point $\tau$ (Sec.~\ref{sec:supp_eval}). \textbf{Left to right, top to bottom:}
precision--recall; F-score versus distance threshold; precision versus threshold; and the
percentile distribution of per-point pred-to-ground-truth distances (lower is better). Increasing
$\lambda_{\text{s}}$ shifts the precision curve upward and the accuracy distribution
downward. The curves shown are geometry only; the corresponding rendering metrics, which peak at
$\lambda_{\text{s}}=10^{-3}$, are reported in Table~\ref{tab:lambda_sweep}.}
\label{fig:lambda_sweep}
\end{figure}

The support-loss weight $\lambda_{\text{s}}$ controls geometric fidelity, and within a moderate
range it improves rendering as well. Although the support loss is a purely geometric constraint
that one would expect to trade against photometric quality, adding it at
$\lambda_{\text{s}}=10^{-3}$ improves all three rendering metrics over the unconstrained
$\lambda_{\text{s}}=0$ configuration (PSNR $24.93\to25.11$, SSIM $0.790\to0.812$, LPIPS
$0.261\to0.230$): constraining neural Gaussians to the measured surface acts as a regularizer on
the photometric fit rather than a cost. Geometry improves far more steeply and over the whole range: precision rises monotonically from $0.211$ to $0.532$, while F-score and Chamfer distance reach their best values at $\lambda_{\text{s}}=10^{-2}$, where the F-score more than doubles ($0.073\to0.181$) and the Chamfer distance falls by more than half ($1.931\to0.805\,\text{m}$) relative to the unconstrained result. This progression is visible across the threshold range
in Figure~\ref{fig:lambda_sweep}: the precision curve shifts upward with
$\lambda_{\text{s}}$ and the pred-to-ground-truth distance distribution shifts downward,
confirming that the effect is consistent across thresholds rather than an artifact of the
operating point. The two objectives diverge beyond $\lambda_{\text{s}}=10^{-3}$: as the
constraint tightens further it begins to override the photometric objective, and rendering
degrades sharply, falling to $23.90\,\text{dB}$ PSNR by $\lambda_{\text{s}}=10^{-1}$ even as
precision keeps climbing. We therefore adopt $\lambda_{\text{s}}=10^{-3}$, the point that
maximizes rendering quality while capturing most of the geometric gain, and note that geometry
can be pushed substantially further at a measured cost in rendering, making the support loss a
controllable operating point on the geometry--rendering trade-off rather than a fixed cost.
 
\begin{table}[t]
\centering
\caption{\textbf{Effect of the support-loss weight} ($\lambda_{\text{s}}$) on the College scene
in the full method, growth enabled, all other settings fixed. $\lambda_{\text{s}}=0$ leaves the
surface-frame offsets unconstrained; increasing it tightens adherence to the measured surface.
$\tau=0.333$ for College; Chamfer in metres, Size in MB. Best per column in \textbf{bold}.}
\label{tab:lambda_sweep}
\setlength{\tabcolsep}{6pt}
\begin{tabular}{l ccc cccc r}
\toprule
& \multicolumn{3}{c}{\textbf{Rendering}} & \multicolumn{4}{c}{\textbf{Geometry} (at $\tau$)} & \textbf{Capacity} \\
\cmidrule(lr){2-4} \cmidrule(lr){5-8} \cmidrule(lr){9-9}
$\lambda_{\text{s}}$ & PSNR$\uparrow$ & SSIM$\uparrow$ & LPIPS$\downarrow$ & Precision$\uparrow$ & Recall$\uparrow$ & F1$\uparrow$ & Ch.$\downarrow$ & Size$\downarrow$ \\
\midrule
$0$              & 24.93 & 0.790 & 0.261 & 0.211 & 0.044 & 0.073 & 1.931 & 191.5 \\
$10^{-4}$        & 25.01 & 0.801 & 0.243 & 0.291 & 0.073 & 0.117 & 1.092 & 192.8 \\
$10^{-3}$ (final)& \textbf{25.11} & \textbf{0.812} & \textbf{0.230} & 0.363 & 0.094 & 0.150 & 0.880 & 194.4 \\
$10^{-2}$        & 24.72 & 0.797 & 0.255 & 0.487 & \textbf{0.111} & \textbf{0.181} & \textbf{0.805} & 194.7 \\
$10^{-1}$        & 23.90 & 0.740 & 0.326 & \textbf{0.532} & 0.064 & 0.114 & 0.938 & \textbf{188.6} \\
\bottomrule
\end{tabular}
\end{table}
 
\subsection{Neural Gaussians per Anchor}
\label{sec:supp_offsets}

The number $k$ of neural Gaussians decoded per anchor directly controls expressive capacity (Table~\ref{tab:offset_sweep}). Increasing $k$ improves both rendering and
alignment at a roughly linear cost in model size: precision rises from $0.368$ to $0.397$ and
Chamfer distance falls from $0.960$ to $0.777\,$m between $k=5$ and $k=20$, while size grows from
$164.5$ to $258.6$\,MB. PSNR peaks at $k=15$, while SSIM and LPIPS improve marginally to $k=20$. We adopt $k=10$, matching the Scaffold-GS default, where
the size cost is balanced against quality. The sweep varies the per-anchor offset budget, not the
number of anchors; adding capacity through anchors rather than offsets is an orthogonal direction
we leave to future work.

\begin{table}[t]
\centering
\small
\setlength{\tabcolsep}{4pt}
\caption{\textbf{Neural Gaussians per anchor} ($k$) on the College scene, all other settings
fixed. $k$ is set per anchor and not adapted during training. $\tau=0.333$\,m; Ch.\ is Chamfer
distance in metres; Size in MB. Best per column in \textbf{bold}.}
\label{tab:offset_sweep}
\begin{tabular}{l ccc cccc r}
\toprule
& \multicolumn{3}{c}{Rendering} & \multicolumn{4}{c}{Geometry (at $\tau$)} & Capacity \\
\cmidrule(lr){2-4}\cmidrule(lr){5-8}\cmidrule(lr){9-9}
$k$ & PSNR$\uparrow$ & SSIM$\uparrow$ & LPIPS$\downarrow$ & P$\uparrow$ & R$\uparrow$ & F1$\uparrow$ & Ch.$\downarrow$ & Size$\downarrow$ \\
\midrule
5            & 24.97 & 0.800 & 0.254 & 0.368 & 0.064 & 0.110 & 0.960 & \textbf{164.5} \\
10 (adopted) & 25.11 & 0.812 & 0.230 & 0.363 & 0.094 & 0.150 & 0.880 & 194.4 \\
15           & \textbf{25.37} & 0.824 & 0.210 & 0.387 & 0.157 & 0.222 & 0.817 & 228.0 \\
20           & 25.33 & \textbf{0.827} & \textbf{0.202} & \textbf{0.397} & \textbf{0.196} & \textbf{0.261} & \textbf{0.777} & 258.6 \\
\bottomrule
\end{tabular}
\end{table}

\end{document}